\documentclass[11pt]{article}

\usepackage[final]{acl}

\usepackage{times}
\usepackage{latexsym}
\usepackage{amsmath}
\usepackage{amssymb}

\usepackage[T1]{fontenc}

\usepackage[utf8]{inputenc}

\usepackage{microtype}

\usepackage{graphicx}
\usepackage{booktabs}
\usepackage{multirow}
\usepackage{booktabs}
\usepackage{multirow}
\usepackage{colortbl}
\usepackage{listings}
\usepackage{cleveref}
\usepackage{cleveref}
\usepackage{placeins}
\usepackage{stfloats}

\lstdefinestyle{aimercode}{
  language=Python,
  basicstyle=\ttfamily\small,
  keywordstyle=\bfseries\color[rgb]{0.30,0.31,0.78},
  commentstyle=\color[rgb]{0.25,0.50,0.50},
  showstringspaces=false,
  keepspaces=true,
  columns=fullflexible,
  breaklines=true,
  frame=none
}

\newcommand{\req}{\textcolor{red!75!black}{$\checkmark$}}
\newcommand{\nreq}{\textcolor{green!50!black}{$\times$}}
\definecolor{aimerrow}{HTML}{E9F2FB}
\newcounter{algorithm}
\renewcommand{\thealgorithm}{\arabic{algorithm}}

\title{AIMER: Calibration-Free Task-Agnostic MoE Expert Pruning}

\author{
  \textbf{Zongfang Liu\textsuperscript{1,2}}
  \quad
  \textbf{Guangyi Chen\textsuperscript{3,4}}
  \quad
  \textbf{Yifan Shen\textsuperscript{4}}
  \quad
  \textbf{Shengkun Tang\textsuperscript{4}}
\\
  \textbf{Huan Wang\textsuperscript{2,\dag}}
  \quad
  \textbf{Xin Yuan\textsuperscript{2,\dag}}
\\
\\
  \normalfont
  \textsuperscript{1}Zhejiang University, Hangzhou, China
  \quad
  \textsuperscript{2}Westlake University, Hangzhou, China
\\
  \textsuperscript{3}Carnegie Mellon University, Pittsburgh, PA, USA
\\
  \textsuperscript{4}Mohamed bin Zayed University of Artificial Intelligence, Abu Dhabi, UAE
\\
  \small{\textsuperscript{\dag}\textbf{Corresponding authors:}
  \href{mailto:wanghuan@westlake.edu.cn}{\texttt{wanghuan@westlake.edu.cn}},
  \href{mailto:xyuan@westlake.edu.cn}{\texttt{xyuan@westlake.edu.cn}}.}
}

\begin{document}
\maketitle
\begin{abstract}
Mixture-of-Experts (MoE) language models increase parameter capacity without proportional per-token computation, yet deployment still requires storing the full expert pool, making expert pruning important for reducing memory and serving overhead. Existing task-agnostic expert-pruning methods are typically calibration-dependent: they estimate expert importance from routing or activation statistics on a calibration set, making pruning decisions sensitive to calibration-data variation while introducing substantial preprocessing cost. We propose AIMER (\textbf{A}bsolute mean over root mean square \textbf{IM}portance for \textbf{E}xpert \textbf{R}anking), a simple calibration-free criterion that identifies more distinct experts by capturing the concentration pattern of expert weights, making it well suited for task-agnostic expert pruning. Across 7B to 47B MoE language models with distinct architectures and 16 diverse benchmarks, AIMER consistently delivers stronger capability balance across diverse tasks than existing calibration-free methods. Surprisingly, AIMER also achieves better balance than strong calibration-based expert-pruning baselines calibrated on the widely used task-agnostic C4 corpus, while requiring only 0.22--2.06 seconds to score all experts.
\end{abstract}

\section{Introduction}
\begin{figure}[!t]
  \centering
  \includegraphics[width=\linewidth]{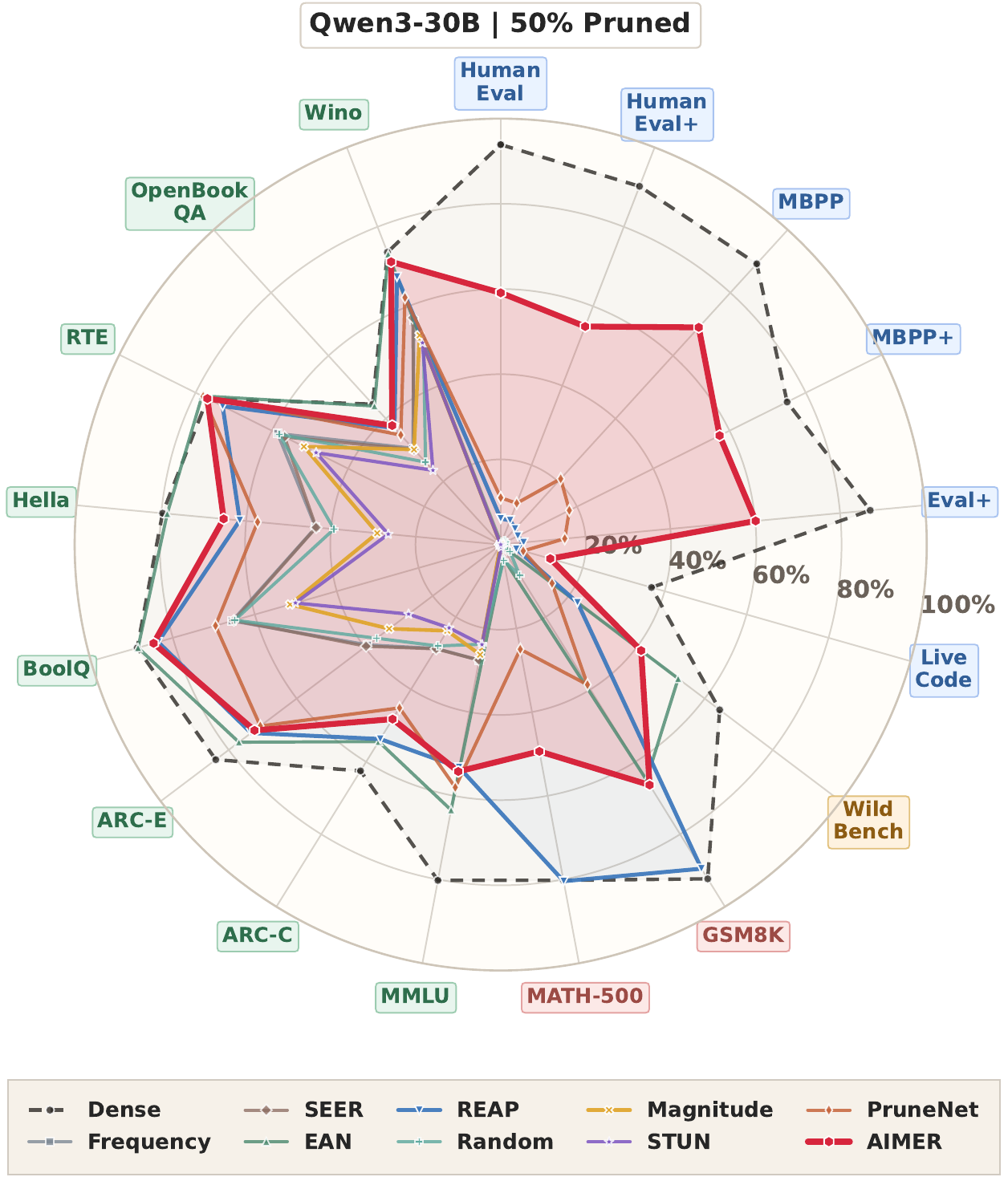}
  \caption{\textbf{Radar plot of Qwen3-30B performance across all benchmarks at 50\% pruning ratio.} The dashed outline denotes the dense model, and each colored trace corresponds to one pruning method. Higher values indicate better task performance on the corresponding benchmark. Among the pruned models, AIMER encloses the largest area and has a capability profile closer to that of the full model. Additional radar plots for the other settings are provided in Appendix~\ref{sec:appendix_radar}.}
  \label{fig:main_radar_qwen50}
\end{figure}

Mixture-of-Experts (MoE) models extend Transformer architectures by replacing the dense feed-forward block with a set of expert FFNs and a router that activates only the top-$k$ experts for each token \cite{vaswani2017attention,shazeer2017outrageously}. This conditional computation paradigm decouples parameter growth from per-token computation, making it possible to scale model capacity without incurring the full inference cost of dense models. As a result, recent MoE large language models \cite{jiang2024mixtral,muennighoff2024olmoe,liu2024deepseek,meta2025llama,baidu2025ernie45, glm5team2026glm5vibecodingagentic,kimiteam2026kimik25visualagentic,qwen3.5} achieve strong performance while maintaining relatively low per-token compute. Despite this advantage, efficient deployment of MoE models remains challenging because all experts must still be stored and managed at inference time, creating substantial memory and serving overhead. Recent routing analyses show that expert usage is often highly imbalanced and that many experts are functionally redundant \cite{huang2024mixture}, motivating a growing body of work on expert-level compression through pruning, merging, and related strategies~\cite{li2023merge,lu2024not,zhang2025diversifying,lee2025stun,lasby2025reap}.

\begin{figure}[t]
  \centering
  \includegraphics[width=1\linewidth]{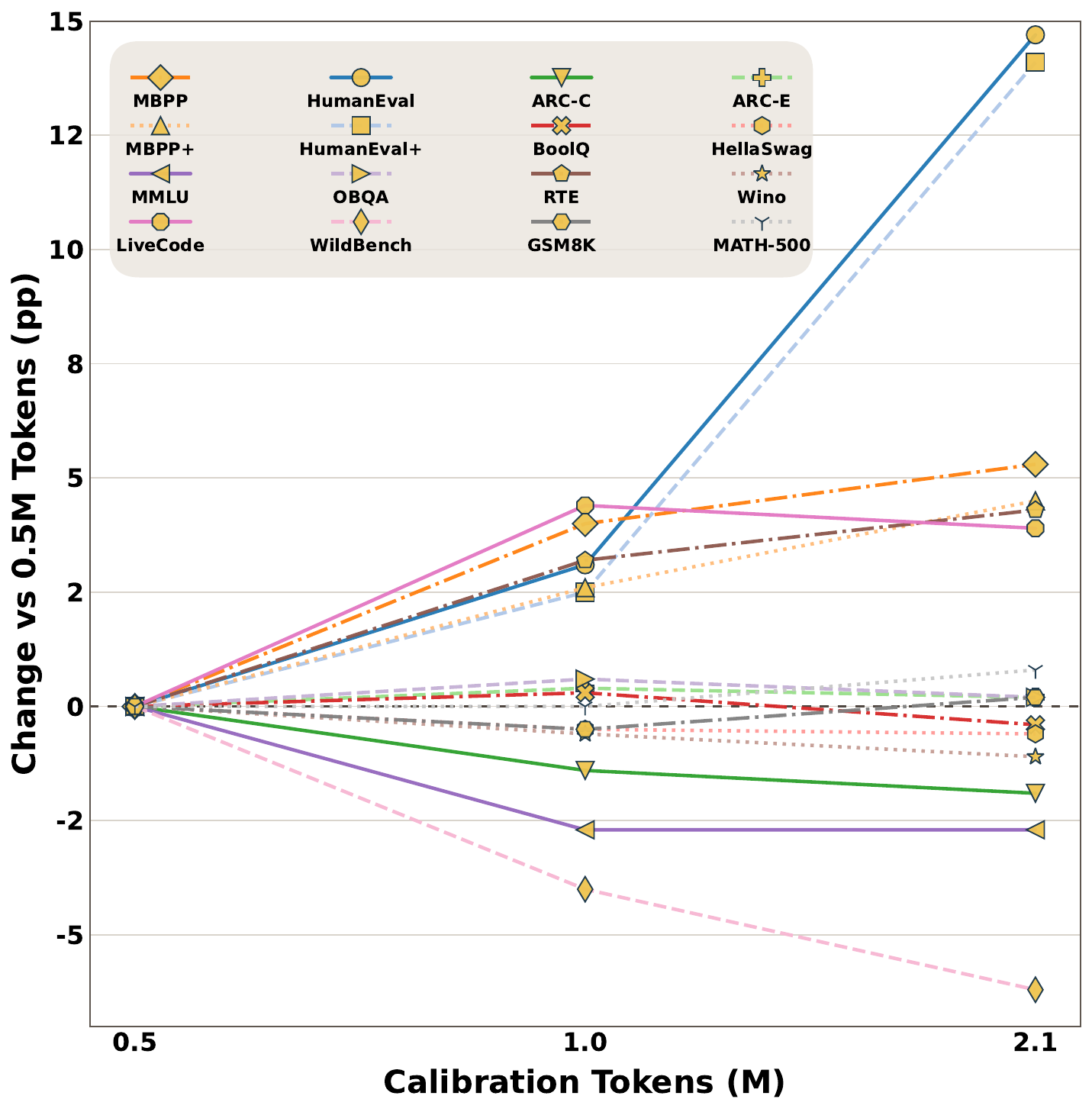}
  \caption{\textbf{Sensitivity of REAP~\cite{lasby2025reap} to calibration set size on Qwen3-30B at 50\% pruning ratio.} We fix the calibration corpus to C4~\cite{allenai_c4_2024} and vary only the size of the calibration set from 0.5M to 2.1M tokens. The horizontal axis reports calibration tokens in millions (M = million tokens), and the vertical axis reports performance change relative to the 0.5M-token setting (pp = percentage points). Half of the benchmarks show significant variation. Some benchmarks improve while others degrade, showing that performance is highly sensitive to calibration set size even with the same corpus.}
  \label{fig:qwen_reap_calibration_size}
\end{figure}

Existing MoE expert pruning methods are typically calibration-dependent, relying on routing or activation statistics collected from a calibration set to rank experts. Prior work has shown that pruning outcomes are highly sensitive to the choice of calibration corpus~\cite{liu2024efficient, zhang2026monereplacingredundantexperts, yang2025moe, liu2026evoesap}. In task-agnostic pruning, C4~\cite{allenai_c4_2024} is widely adopted as a general-domain calibration corpus~\cite{frantar2023sparsegpt,bandari2024c4,xia2024sheared,ling2024slimgpt}. Yet using a standard general-domain corpus does not remove the sensitivity to calibration data. As shown in \Cref{fig:qwen_reap_calibration_size}, even with the corpus fixed to C4, varying only the number of calibration tokens already leads to substantially different pruning outcomes. This observation suggests that expert ranking remains inherently tied to the sampled distribution, introducing an irreducible bias toward experts that are useful for the sampled data rather than those that generalize across tasks. This raises a natural question for task-agnostic MoE
pruning: \textit{Can removable experts be identified without relying on calibration data?}

In this work, we propose AIMER, a calibration-free expert-pruning criterion that scores each expert by the concentration pattern of its weights, computed as the mean absolute value normalized by the root-mean-square value. Empirically, through CKA~\cite{kornblith2019similarityneuralnetworkrepresentations} analysis we find that AIMER tends to retain more distinct experts, which helps explain why it is well suited for task-agnostic pruning: the retained expert pool should preserve diverse functionality rather than specialize to a particular capability. We evaluate AIMER on five representative MoE model families spanning distinct architectures and scales from 7B to 47B parameters. We compare against both calibration-free baselines and calibration-based expert-pruning baselines calibrated on C4 across 16 benchmarks covering coding, creative writing, mathematical reasoning, and multiple-choice question answering. Across these settings, AIMER achieves the best average rank on four of the five model families and ties for best on the fifth, indicating the strongest overall capability balance, while eliminating most of the calibration overhead: expert scoring takes only 0.22--2.06 seconds, compared with 0.75--2.96 hours for calibration-based REAP on C4. An illustrative example of this balanced capability preservation is shown for Qwen3 at 50\% pruning in \Cref{fig:main_radar_qwen50}. The accompanying implementation is publicly available.\footnote{\url{https://github.com/ZongfangLiu/AIMER}}

\noindent\textbf{Our main contributions are as follows:}
\begin{itemize}
    \item We highlight calibration dependence as a central limitation of task-agnostic MoE expert pruning, and provide evidence that pruning outcomes can remain sensitive even when calibration data is sampled from the same corpus.
    \item We propose AIMER, a simple calibration-free criterion for task-agnostic expert ranking in MoE language models that captures the concentration pattern of expert weights, and empirically show that it retains a more distinct set of experts after pruning.
    \item We show that across five MoE model families and 16 benchmarks, AIMER delivers stronger overall capability balance than strong expert-pruning baselines while reducing expert-scoring time from hours to seconds.
\end{itemize}

\section{Related Work}

\subsection{Expert Pruning for MoE language models}
Expert pruning in MoE models was first explored in task-adaptive settings. \citet{chen2022task} progressively drop non-professional experts for a downstream task, showing that substantial redundancy can be removed after fine-tuning. In multilingual machine translation, \citet{koishekenov2023memory} prune language-specific experts to improve memory efficiency at deployment time. These studies establish that MoE models often contain significant redundancy, but their pruning strategies are limited to task-specific settings. More recent work studies task-agnostic compression for modern MoE language models. NAEE~\cite{lu2024not} selects retained experts by minimizing Frobenius-norm reconstruction error between the original and pruned layer outputs, and EEP~\cite{liu2024efficient} uses a gradient-free evolutionary strategy to search for effective expert subsets. \citet{zhang2025diversifying} prune groups of similar experts to preserve diversity among the retained experts, and STUN~\cite{lee2025stun} clusters experts first to prune redundant ones and then applies unstructured pruning on remaining experts. Seer-MoE~\cite{muzio2024seer} scores experts with heavy-hitters counting, either through hard activation counts or, in its soft-counting variant, by accumulating router softmax probabilities as weighted expert frequencies. REAP~\cite{lasby2025reap} estimates expert importance from router-weighted activations in a one-shot pruning pipeline. \citet{jaiswal2025finding} benchmark 16 expert-importance criteria for expert dropping and report expert activation norm (EAN) as the strongest criterion among them. MoNE~\cite{zhang2026monereplacingredundantexperts} identifies redundant experts using routing-frequency-weighted activation variance and replaces them with lightweight novice experts.

\subsection{Expert Merging for MoE language models}
MEO~\cite{he2023merging} performs merging online at inference time, constructing a merged expert for each token as a router-score-weighted combination of the activated experts. Subsequent methods instead adopt offline merging. MC-SMoE~\cite{li2023merge} first aligns neurons across experts and then merges routing-based groups using activation-frequency-weighted averaging. HC-SMoE~\cite{chen2024retraining} builds a hierarchical clustering of experts based on output similarity and merges experts within each cluster by frequency-weighted averaging. DERN~\cite{zhou2025dropping} goes beyond whole-expert merging: after pruning redundant experts, it decomposes them into neuron-level segments and reallocates those segments to compatible retained experts. Despite differences in scoring criteria and compression pipelines, most expert pruning and merging methods still rely on routing statistics, activation measurements, or calibration-set evaluations to decide which experts to remove or merge.

\subsection{Calibration-Free Model Pruning}
Calibration-free pruning studies whether removable structure can be identified from the pretrained model itself, without collecting activations on a held-out calibration set. Early work in dense networks approached this question through explicit weight-space redundancy. \citet{srinivas2015data} show that pruning can be performed one neuron at a time by identifying similar neurons and removing those whose contribution can be absorbed by the remaining weights. \citet{mussay2019data} develop this perspective further by casting pruning as a coreset construction problem, selecting a small weighted subset of neurons that approximates the original layer with provable guarantees for arbitrary future inputs. 
Subsequent work shifts from neuron selection toward reconstruction-based compensation. RED++~\cite{yvinec2022red++} exploits redundancies in neuron weights through data-free hashing and removes input-wise redundant operations via input splitting and output merging, showing that structured pruning can be carried out without access to data. More recently, \citet{sengupta2025you} bring the calibration-free view to dense LLM compression with PruneNet, which reformulates pruning as policy learning over intrinsic model properties instead of relying on calibration examples. Taken together, these studies suggest that informative compression signals can often be derived directly from pretrained weights themselves, without relying on external calibration sets. Yet calibration-free expert-level pruning for MoE models remains largely unexplored.

\section{Preliminary}
\noindent\textbf{Mixture-of-Experts.} An MoE layer replaces the dense feed-forward network in a standard Transformer block with a collection of $n$ expert FFNs $\{E_i\}_{i=1}^{n}$ and a router that activates only a small subset of them for each token~\cite{shazeer2017outrageously}. This design preserves large model capacity while avoiding the cost of evaluating every expert on every input. Given a token representation $h\in\mathbb{R}^{d}$, the router first computes an expert logit vector
\[
z = \mathbf{W}_r h \in \mathbb{R}^{n},
\]
where $\mathbf{W}_r \in \mathbb{R}^{n\times d}$ is the router projection matrix. Let $\mathcal{E}(h)=\mathrm{TopK}(z,k)$ denote the indices of the top-$k$ logits, with $k\ll n$. The router then normalizes scores only over these selected experts, yielding sparse gating weights
\begin{equation}
g_i(h)=
\begin{cases}
\dfrac{\exp(z_i)}{\sum_{j\in \mathcal{E}(h)}\exp(z_j)}, & i\in \mathcal{E}(h),\\[6pt]
0, & i\notin \mathcal{E}(h),
\end{cases}
\label{eq:topk_gate}
\end{equation}
The layer output is the weighted combination of the activated experts, where $A_i(h)=E_i(h)$ denotes the output activation produced by expert $E_i$:
\begin{equation}
y(h) \;=\; \sum_{i\in \mathcal{E}(h)} g_i(h)\,A_i(h).
\label{eq:moe_output}
\end{equation}
Since $g(h)$ has only $k$ nonzero entries, each token is routed to just a few experts. As a result, the computation per-token scales with $k$ rather than $n$, while the total parameter capacity still grows with the full expert pool~\cite{fedus2022switch}. 

\begin{figure}[!t]
\refstepcounter{algorithm}\label{alg:aimer}
\small
\hrule
\vspace{0.4em}
\noindent\textbf{Algorithm \thealgorithm} PyTorch-style expert ranking with AIMER
\vspace{0.4em}
\hrule
\vspace{0.3em}
\begin{lstlisting}[style=aimercode]
def aimer_rank(layer):
    scores = []
    num_experts = layer.gate.weight.shape[0]

    for e in range(num_experts):
        gate, up, down = get_proj_weights(layer, e)
        abs_sum = gate.abs().sum() + up.abs().sum() + down.abs().sum()
        numel = gate.numel() + up.numel() + down.numel()
        l2_sq = gate.square().sum() + up.square().sum() + down.square().sum()
        score = (abs_sum / numel) / torch.sqrt(l2_sq / numel)
        scores.append(score)

    scores = torch.stack(scores)
    _, sorted_idx = torch.sort(scores, descending=True)
    return sorted_idx
\end{lstlisting}
\vspace{0.2em}
\hrule
\end{figure}

\section{Methodology}
\begin{figure*}[!t]
  \centering
  \includegraphics[width=\textwidth]{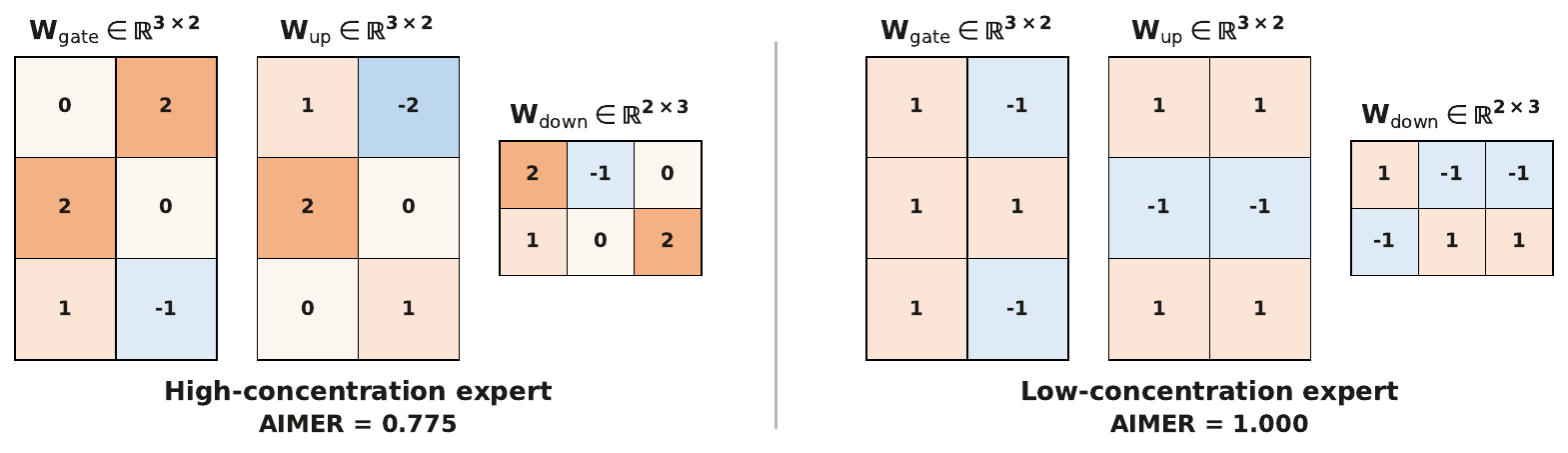}
  \caption{\textbf{A schematic example of AIMER scoring.} We consider two experts with the same architecture but different absolute-weight patterns. When the weight mass is concentrated in a few entries (left), the RMS is larger relative to the absolute mean, producing a lower AIMER score. When all weights have equal magnitude (right), the absolute mean equals the RMS and the score reaches its maximum value of 1.}
  \label{fig:aimer_absmean_over_rms_example}
\end{figure*}

\subsection{Problem Formulation}
The goal of task-agnostic expert pruning is to reduce the expert pool while preserving the model's overall capability across diverse tasks, rather than leaning to any specific one. Consider an MoE language model with $L$ layers, where layer $\ell$ contains $n_\ell$ experts $\{E_i^{(\ell)}\}_{i=1}^{n_\ell}$. Given a target pruning ratio $\rho$, every layer removes the same fraction of experts. 
In this setup, the pruning problem reduces to a problem of expert ranking within the layer. For each layer $\ell$, we assign each expert $E_i^{(\ell)}$ a scalar score $s_i^{(\ell)}$, sort experts within that layer by this score, and prune the $\rho  n_\ell$ experts judged as more redundant. After the pruning decision is made, we remove the selected expert FFN parameters together with the corresponding rows in the router matrix, and retain the same top-$k$ routing rule over the remaining experts. The remaining question is therefore how to define the within-layer score $s_i^{(\ell)}$. We answer this with AIMER, a criterion that assigns each expert a score computed directly from its parameters.

\subsection{Proposed Method}

\noindent \textbf{The AIMER criterion.} For one expert, let $d$ denote the input dimension and $m$ denote the hidden dimension. Then $\mathbf{W}_{\mathrm{gate}}, \mathbf{W}_{\mathrm{up}} \in \mathbb{R}^{m\times d}$ and $\mathbf{W}_{\mathrm{down}} \in \mathbb{R}^{d\times m}$ are the gate, up, and down projection matrices of that expert. We define
\begin{equation}
\begin{aligned}
N &= N_{\mathrm{gate}} + N_{\mathrm{up}} + N_{\mathrm{down}}, \\
P &= \left\|\mathbf{W}_{\mathrm{gate}}\right\|_1
   + \left\|\mathbf{W}_{\mathrm{up}}\right\|_1
   + \left\|\mathbf{W}_{\mathrm{down}}\right\|_1, \\
Q &= \left\|\mathbf{W}_{\mathrm{gate}}\right\|_F^2
   + \left\|\mathbf{W}_{\mathrm{up}}\right\|_F^2
   + \left\|\mathbf{W}_{\mathrm{down}}\right\|_F^2,
\end{aligned}
\label{eq:aimer_components}
\end{equation}
where $N_{\mathrm{gate}}=N_{\mathrm{up}}=N_{\mathrm{down}}=md$ are the numbers of parameters in the three matrices. AIMER is then
\begin{equation}
\mathrm{AIMER}
=
\frac{P/N}{\sqrt{Q/N}}
=
\frac{P}{\sqrt{NQ}}.
\label{eq:aimer_score}
\end{equation}
and we prune experts with larger AIMER scores. The layer-wise procedure is shown in Algorithm~\ref{alg:aimer}.

If we flatten and concatenate the three projection matrices of an expert into a single vector $\mathbf{w}\in\mathbb{R}^N$, then
\begin{equation}
\begin{aligned}
\mathrm{AIMER}(\mathbf{w})
&=
\frac{\|\mathbf{w}\|_1}{\sqrt{N}\,\|\mathbf{w}\|_2}.
\end{aligned}
\label{eq:aimer_vector}
\end{equation}
This vector form is algebraically equivalent to \Cref{eq:aimer_score} and is in the same spirit as the Hoyer metric~\cite{hoyer2004non}:
\begin{equation}
\mathrm{Hoyer}(\mathbf{w})
=
\frac{\sqrt{N}-\|\mathbf{w}\|_1/\|\mathbf{w}\|_2}{\sqrt{N}-1}.
\end{equation}
That is, both AIMER and the Hoyer metric are functions of the same underlying $\ell_1/\ell_2$ ratio.
Hoyer and DeepHoyer~\cite{yang2020deephoyer} use the $\ell_1/\ell_2$ norm as a training-time regularizer, typically on vectors or channels within a weight matrix. In contrast, AIMER uses the same underlying quantity only after training, as a calibration-free weight-only criterion that treats the entire expert as a single vector and assigns it a scalar ranking score.

\noindent \textbf{Basic properties.}
Let $\mathbf{w}\in\mathbb{R}^{N}$ denote the flattened parameter vector of one expert. AIMER has two useful properties for expert comparison. First, it is \emph{scale-invariant}: for any nonzero scalar $c$, $\mathrm{AIMER}(c\mathbf{w})=\mathrm{AIMER}(\mathbf{w})$. This is desirable for ranking because the goal is to compare experts by the relative pattern of their parameters, rather than by their overall scale. Second, it is \emph{bounded}: combining $\|\mathbf{w}\|_1\ge \|\mathbf{w}\|_2$ with Cauchy--Schwarz gives
\begin{equation}
\frac{1}{\sqrt{N}}
\;\le\;
\mathrm{AIMER}(\mathbf{w})
\;\le\;
1.
\label{eq:aimer_bounds}
\end{equation}
The upper bound is attained when all entries have equal absolute value, whereas the lower bound is attained when only one element is nonzero. Since $N$ is fixed within a layer, all experts in that layer share the same bounded score range, making AIMER directly comparable across experts. The factor $1/\sqrt{N}$ therefore does not change the within-layer ranking and is kept only for interpretability.

\begin{figure*}[!t]
  \centering
  \includegraphics[width=\textwidth]{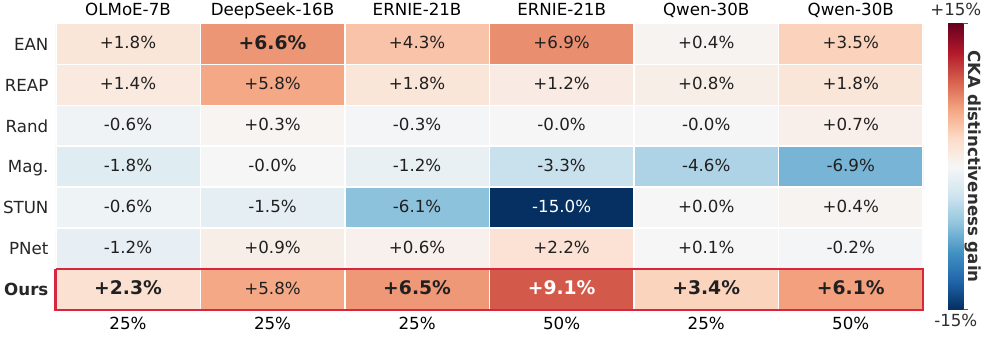}
  \caption{\textbf{Change in expert distinctiveness after pruning.}
For each model and pruning ratio, we use 4096 C4 tokens to compute pairwise CKA similarities among expert outputs within each layer. Each cell reports the layer-averaged relative decrease in mean CKA after pruning,
$(\mathrm{CKA}_{\mathrm{all}}-\mathrm{CKA}_{\mathrm{kept}})/\mathrm{CKA}_{\mathrm{all}}$.
Positive values indicate that the retained experts are less similar to one another than the original expert pool, i.e., more functionally distinct; negative values indicate increased redundancy among retained experts. AIMER (Ours) yields positive distinctiveness gains in all shown settings and is strongest or near-strongest in most of them.}
  \label{fig:cka_distinctness_gain}
\end{figure*}

\noindent \textbf{Explanation.}
AIMER measures the concentration pattern of an expert's weights through a normalized $\ell_1/\ell_2$ ratio. As illustrated in \Cref{fig:aimer_absmean_over_rms_example},  a larger AIMER score means that the weight mass is more evenly distributed across parameters, whereas a smaller score means that the mass is concentrated in fewer large parameters. Empirically, \Cref{fig:cka_distinctness_gain} shows that AIMER yields positive CKA distinctiveness gains in all settings and is the strongest method in all settings except DeepSeek-16B, where it ranks second. This trend also holds on code-domain inputs (Appendix~\ref{sec:appendix_code_cka}), indicating that the observed gains are not specific to C4. These results suggest that AIMER tends to remove redundant experts while preserving diverse expert transformations, which is desirable for task-agnostic pruning, where the underlying distribution is unknown and thus no target capability is preferred.

\section{Experimental Results}

\subsection{Experimental Setup}
\noindent \textbf{Models and baselines.}
We evaluate AIMER on five representative MoE LLMs spanning different architectures and scales: OLMoE-1B-7B-0125-Instruct~\cite{muennighoff2024olmoe}, DeepSeek-V2-Lite-Chat~\cite{deepseekv2}, ERNIE-4.5-21B-A3B-PT~\cite{baidu2025ernie45}, Qwen3-30B-A3B-Instruct-2507~\cite{yang2025qwen3}, and Mixtral-8x7B-Instruct-v0.1~\cite{jiang2024mixtral}. Their architectural details are summarized in Appendix~\ref{sec:appendix_model_architecture}. We report the main results at 25\% pruning for OLMoE and DeepSeek, and at 50\% pruning for ERNIE, Qwen3, and Mixtral; full per-benchmark results and additional pruning settings are provided in Appendix~\ref{sec:appendix}. We compare AIMER against four calibration-based expert-pruning baselines: REAP~\cite{lasby2025reap}, Expert Activation Norm (EAN)~\cite{jaiswal2025finding}, Frequency, and SEER soft counting~\cite{muzio2024seer}. Table~\ref{tab:metric_comparison} summarizes the extra signals required by these calibration-based criteria, and Appendix~\ref{sec:appendix_metric_scores} gives the corresponding score definitions. For a controlled comparison, all calibration-based baselines use the same 4.2M-token calibration set sampled from C4~\cite{allenai_c4_2024}. We also compare with four calibration-free baselines: Random, Magnitude, STUN~\cite{lee2025stun}, and PruneNet~\cite{sengupta2025you}.

\noindent \textbf{Evaluation suite.}
To assess whether pruning decisions generalize beyond a narrow task distribution, we evaluate the pruned models in the zero-shot setting on 16 benchmarks covering both discriminative reasoning and open-ended generation. For multiple-choice evaluation, we report AI2 Reasoning Challenge (ARC-C/ARC-E)~\cite{clark2018think}, BoolQ~\cite{clark2019boolq}, HellaSwag~\cite{zellers2019hellaswag}, MMLU~\cite{hendrycks2020measuring}, OpenBookQA (OBQA)~\cite{mihaylov2018can}, Recognizing Textual Entailment (RTE)~\cite{bentivogli2009fifth}, and WinoGrande (WinoG.)~\cite{sakaguchi2021winogrande}, all implemented with lm-eval-harness~\cite{gao2021framework}. For open-ended generation, we evaluate code generation on EvalPlus~\cite{liu2023your} and 182 LiveCodeBench~\cite{jain2024livecodebench} problems collected between January and April 2025; mathematical reasoning on GSM8K~\cite{cobbe2021training} and MATH-500~\cite{hendrycks2021measuring} using EvalScope~\cite{evalscope_2024}; and creative writing on 146 prompts sampled from WildBench~\cite{lin2024wildbench}. For WildBench, we use \texttt{gpt-oss-120b}~\cite{openai2025gptoss120bgptoss20bmodel} as the judge. Overall, this suite spans multiple-choice QA, coding, math, and creative generation, allowing us to assess pruning robustness across substantially different capabilities.

\begin{table}[!t]
  \centering
  \setlength{\tabcolsep}{5pt}
  \renewcommand{\arraystretch}{1.05}
  \resizebox{0.80\linewidth}{!}{%
  \begin{tabular}{l c c c}
    \toprule
    \textbf{Method} & \textbf{Data} & \textbf{Act.} & \textbf{Route} \\
    \midrule
    Frequency & \req & \nreq & \req \\
    SEER & \req & \nreq & \req \\
    EAN & \req & \req & \nreq \\
    REAP & \req & \req & \req \\
    \textbf{AIMER (Ours)} & \nreq & \nreq & \nreq \\
    \bottomrule
  \end{tabular}
  }
  \caption{\textbf{Comparison of statistics requirements for AIMER and calibration-based baselines.} Data = calibration set, Act. = expert activations, and Route = router weights. AIMER ranks experts from pretrained weights alone, so none of these extra signals are required. Red marks required; green marks not required.}
  \label{tab:metric_comparison}
\end{table}

\noindent \textbf{Protocol and hardware.}
Scores are computed once per model and criterion. Only routed experts are pruned; shared experts and dense FFN layers, where present, are preserved, and no recovery training is performed. All evaluations are conducted in the zero-shot setting to isolate the effect of expert pruning from task-specific adaptation. For open-ended generation, we use deterministic decoding with \texttt{do\_sample=False}; when an explicit temperature parameter is exposed, we set \texttt{temperature=0}. In practice, this corresponds to greedy, non-sampled generation, improving reproducibility across pruning methods. Experiments are conducted on NVIDIA L40S 48GB GPUs.

\FloatBarrier

\begin{table*}[tbp]
  \centering
  \resizebox{0.90\textwidth}{!}{%
  \begin{tabular}{l l l | c c c | c | c c c | c | c}
    \toprule
    \multicolumn{3}{c|}{} & \multicolumn{3}{c|}{\textbf{Coding}} & \multicolumn{1}{c|}{\textbf{Creative Writing}} & \multicolumn{3}{c|}{\textbf{Math}} & \multicolumn{1}{c|}{\textbf{MC}} & \multicolumn{1}{c}{\textbf{Rank}} \\
    \textbf{Model} & \textbf{Ratio} & \textbf{Method} & \textbf{Eval+} & \textbf{LiveCode} & \textbf{Code Avg} & \textbf{WildBench} & \textbf{GSM8K} & \textbf{MATH-500} & \textbf{Math Avg} & \textbf{MC Avg} & \textbf{Rank Avg} \\
    \midrule
    \multirow{10}{*}{OLMoE} & 0\% & Full & 0.341 & 0.033 & 0.187 & 0.444 & 0.682 & 0.222 & 0.452 & 0.653 & N/A \\
    \cmidrule(lr){2-12}
     & \multirow{9}{*}{25\%} & Frequency & 0.000 & 0.000 & 0.000 & 0.127 & 0.033 & 0.024 & 0.028 & 0.560 & 6.67 \\
     &  & SEER & 0.000 & 0.000 & 0.000 & 0.141 & 0.037 & 0.012 & 0.025 & 0.564 & 6.00 \\
     &  & EAN & 0.000 & 0.000 & 0.000 & 0.205 & 0.115 & 0.006 & 0.061 & 0.577 & 5.50 \\
     &  & REAP & 0.014 & 0.000 & 0.007 & 0.260 & \underline{0.202} & \textbf{0.056} & \underline{0.129} & \underline{0.588} & \underline{2.92} \\
     &  & Random & \textbf{0.112} & 0.000 & \textbf{0.056} & \textbf{0.297} & 0.140 & 0.034 & 0.087 & 0.562 & 3.50 \\
     &  & Magnitude & 0.000 & 0.000 & 0.000 & 0.020 & 0.072 & 0.004 & 0.038 & 0.478 & 7.58 \\
     &  & STUN & 0.000 & 0.000 & 0.000 & 0.089 & 0.072 & 0.008 & 0.040 & 0.497 & 6.92 \\
     &  & PruneNet & \underline{0.102} & 0.000 & \underline{0.051} & \underline{0.273} & 0.162 & \underline{0.044} & 0.103 & 0.563 & 3.33 \\
     &  & \cellcolor{aimerrow}AIMER (Ours) & \cellcolor{aimerrow}0.026 & \cellcolor{aimerrow}0.000 & \cellcolor{aimerrow}0.013 & \cellcolor{aimerrow}0.236 & \cellcolor{aimerrow}\textbf{0.221} & \cellcolor{aimerrow}\textbf{0.056} & \cellcolor{aimerrow}\textbf{0.139} & \cellcolor{aimerrow}\textbf{0.600} & \cellcolor{aimerrow}\textbf{2.58} \\
    \midrule[\heavyrulewidth]
    \multirow{10}{*}{ERNIE} & 0\% & Full & 0.867 & 0.247 & 0.557 & 0.479 & 0.829 & 0.780 & 0.804 & 0.721 & N/A \\
    \cmidrule(lr){2-12}
     & \multirow{9}{*}{50\%} & Frequency & 0.007 & 0.000 & 0.004 & 0.164 & 0.051 & 0.018 & 0.034 & 0.557 & 7.58 \\
     &  & SEER & 0.003 & 0.000 & 0.002 & 0.174 & 0.106 & 0.020 & 0.063 & 0.543 & 7.08 \\
     &  & EAN & 0.009 & 0.000 & 0.004 & \textbf{0.253} & 0.065 & 0.026 & 0.046 & \textbf{0.605} & 5.25 \\
     &  & REAP & 0.024 & 0.000 & 0.012 & 0.199 & \underline{0.567} & 0.178 & \underline{0.373} & 0.592 & 4.08 \\
     &  & Random & 0.220 & 0.011 & 0.116 & 0.158 & 0.420 & \underline{0.198} & 0.309 & 0.584 & 4.17 \\
     &  & Magnitude & \underline{0.243} & \underline{0.049} & \underline{0.146} & 0.077 & 0.351 & 0.130 & 0.240 & 0.503 & 5.00 \\
     &  & STUN & 0.117 & 0.005 & 0.061 & 0.044 & 0.103 & 0.066 & 0.085 & 0.487 & 6.83 \\
     &  & PruneNet & 0.199 & 0.022 & 0.111 & 0.180 & 0.461 & 0.154 & 0.307 & 0.586 & \underline{3.67} \\
     &  & \cellcolor{aimerrow}AIMER (Ours) & \cellcolor{aimerrow}\textbf{0.254} & \cellcolor{aimerrow}\textbf{0.071} & \cellcolor{aimerrow}\textbf{0.163} & \cellcolor{aimerrow}\underline{0.217} & \cellcolor{aimerrow}\textbf{0.650} & \cellcolor{aimerrow}\textbf{0.348} & \cellcolor{aimerrow}\textbf{0.499} & \cellcolor{aimerrow}\underline{0.598} & \cellcolor{aimerrow}\textbf{1.33} \\
    \bottomrule
  \end{tabular}%
  }
  \caption{\textbf{Expert-pruning results on OLMoE-1B-7B-0125-Instruct (OLMoE) and ERNIE-4.5-21B-A3B-PT (ERNIE).} AIMER is compared with calibration-based baselines calibrated on C4 and with calibration-free baselines. It achieves the best rank average in both settings, indicating the strongest overall capability balance.}
  \label{tab:main_olmoe_ernie_summary}
\end{table*}

\FloatBarrier

\subsection{Main Results}
The main results are reported in \Cref{tab:main_olmoe_ernie_summary,tab:main_deepseek_qwen_mixtral_summary}. Across five model families and 16 benchmarks, AIMER delivers the strongest or tied-strongest overall capability balance among calibration-free methods. It achieves the best rank average on OLMoE-7B, DeepSeek-16B, ERNIE-21B, and Qwen3-30B, and ties STUN for the best rank average on Mixtral-8x7B. More surprisingly, AIMER also provides better overall balance than strong calibration-based expert-pruning baselines calibrated on C4, despite requiring no calibration set, activations, or router statistics. The category-level results show the same pattern: AIMER is strongest on math and multiple-choice evaluation for OLMoE, strongest on math and second-best on coding for DeepSeek, strongest on coding and math for ERNIE, and strongest on coding for both Qwen3 and Mixtral. On Qwen3 at 50\% pruning, for example, AIMER reaches a 36.1\% code average, compared with 4.6\% for REAP and 0\% for Frequency, SEER, and EAN. Although other methods can still win individual capability groups, such as REAP on Qwen3 math and STUN on part of Mixtral, AIMER provides the most balanced overall profile. This is exactly the behavior we want in task-agnostic pruning: the retained experts should support broad general ability rather than reflect the preferences of a particular calibration sample. Appendix~\ref{sec:appendix_mixing_data} studies how REAP changes under different calibration choices, including Evol-CodeAlpaca-v1~\cite{luo2023wizardcoder} and mixed data.

\begin{table*}[tbp]
  \centering
  \resizebox{0.90\textwidth}{!}{%
  \begin{tabular}{l l l | c c c | c | c c c | c | c}
    \toprule
    \multicolumn{3}{c|}{} & \multicolumn{3}{c|}{\textbf{Coding}} & \multicolumn{1}{c|}{\textbf{Creative Writing}} & \multicolumn{3}{c|}{\textbf{Math}} & \multicolumn{1}{c|}{\textbf{MC}} & \multicolumn{1}{c}{\textbf{Rank}} \\
    \textbf{Model} & \textbf{Ratio} & \textbf{Method} & \textbf{Eval+} & \textbf{LiveCode} & \textbf{Code Avg} & \textbf{WildBench} & \textbf{GSM8K} & \textbf{MATH-500} & \textbf{Math Avg} & \textbf{MC Avg} & \textbf{Rank Avg} \\
    \midrule
    \multirow{10}{*}{DeepSeek} & 0\% & Full & 0.549 & 0.104 & 0.327 & 0.418 & 0.610 & 0.298 & 0.454 & 0.678 & N/A \\
    \cmidrule(lr){2-12}
     & \multirow{9}{*}{25\%} & Frequency & 0.000 & 0.000 & 0.000 & \underline{0.291} & 0.023 & 0.012 & 0.017 & 0.602 & 6.17 \\
     &  & SEER & 0.000 & 0.000 & 0.000 & 0.155 & 0.034 & 0.016 & 0.025 & 0.602 & 6.33 \\
     &  & EAN & 0.000 & 0.000 & 0.000 & \textbf{0.307} & \underline{0.290} & 0.018 & 0.154 & \underline{0.625} & 4.00 \\
     &  & REAP & 0.001 & 0.000 & 0.001 & 0.152 & 0.242 & \underline{0.046} & 0.144 & \textbf{0.640} & 3.92 \\
     &  & Random & 0.000 & 0.000 & 0.000 & 0.050 & 0.081 & 0.018 & 0.050 & 0.521 & 6.67 \\
     &  & Magnitude & 0.000 & 0.000 & 0.000 & 0.003 & 0.008 & 0.002 & 0.005 & 0.419 & 8.25 \\
     &  & STUN & 0.065 & \textbf{0.011} & 0.038 & 0.137 & 0.077 & 0.030 & 0.053 & 0.554 & 4.75 \\
     &  & PruneNet & \textbf{0.185} & \underline{0.005} & \textbf{0.095} & 0.221 & 0.281 & 0.038 & \underline{0.160} & 0.604 & \underline{3.00} \\
     &  & \cellcolor{aimerrow}AIMER (Ours) & \cellcolor{aimerrow}\underline{0.163} & \cellcolor{aimerrow}\textbf{0.011} & \cellcolor{aimerrow}\underline{0.087} & \cellcolor{aimerrow}0.246 & \cellcolor{aimerrow}\textbf{0.493} & \cellcolor{aimerrow}\textbf{0.134} & \cellcolor{aimerrow}\textbf{0.313} & \cellcolor{aimerrow}0.621 & \cellcolor{aimerrow}\textbf{1.92} \\
    \midrule[\heavyrulewidth]
    \multirow{10}{*}{Qwen3} & 0\% & Full & 0.871 & 0.368 & 0.619 & 0.644 & 0.923 & 0.802 & 0.862 & 0.737 & N/A \\
    \cmidrule(lr){2-12}
     & \multirow{9}{*}{50\%} & Frequency & 0.000 & 0.000 & 0.000 & 0.015 & 0.000 & 0.000 & 0.000 & 0.437 & 7.00 \\
     &  & SEER & 0.000 & 0.000 & 0.000 & 0.018 & 0.000 & 0.000 & 0.000 & 0.436 & 7.00 \\
     &  & EAN & 0.000 & 0.000 & 0.000 & \textbf{0.522} & 0.650 & 0.032 & 0.341 & \textbf{0.697} & 4.00 \\
     &  & REAP & 0.054 & 0.038 & 0.046 & 0.226 & \textbf{0.895} & \textbf{0.804} & \textbf{0.849} & 0.628 & \underline{2.33} \\
     &  & Random & 0.010 & 0.005 & 0.007 & 0.027 & 0.085 & 0.038 & 0.061 & 0.409 & 4.83 \\
     &  & Magnitude & 0.000 & 0.000 & 0.000 & 0.000 & 0.000 & 0.000 & 0.000 & 0.372 & 7.75 \\
     &  & STUN & 0.000 & 0.000 & 0.000 & 0.000 & 0.002 & 0.006 & 0.004 & 0.342 & 7.25 \\
     &  & PruneNet & \underline{0.151} & \underline{0.055} & \underline{0.103} & 0.151 & 0.387 & 0.250 & 0.319 & 0.595 & 3.17 \\
     &  & \cellcolor{aimerrow}AIMER (Ours) & \cellcolor{aimerrow}\textbf{0.601} & \cellcolor{aimerrow}\textbf{0.121} & \cellcolor{aimerrow}\textbf{0.361} & \cellcolor{aimerrow}\underline{0.413} & \cellcolor{aimerrow}\underline{0.664} & \cellcolor{aimerrow}\underline{0.494} & \cellcolor{aimerrow}\underline{0.579} & \cellcolor{aimerrow}\underline{0.638} & \cellcolor{aimerrow}\textbf{1.67} \\
    \midrule[\heavyrulewidth]
    \multirow{10}{*}{Mixtral} & 0\% & Full & 0.512 & 0.148 & 0.330 & 0.427 & 0.661 & 0.294 & 0.478 & 0.737 & N/A \\
    \cmidrule(lr){2-12}
     & \multirow{9}{*}{50\%} & Frequency & 0.033 & 0.000 & 0.017 & 0.139 & 0.054 & 0.020 & 0.037 & 0.520 & 8.25 \\
     &  & SEER & 0.035 & 0.000 & 0.018 & 0.130 & 0.070 & 0.010 & 0.040 & 0.523 & 8.08 \\
     &  & EAN & 0.152 & 0.000 & 0.076 & 0.188 & 0.205 & 0.038 & 0.121 & \textbf{0.674} & 4.67 \\
     &  & REAP & 0.166 & 0.011 & 0.089 & 0.244 & \textbf{0.347} & \underline{0.054} & \textbf{0.201} & 0.670 & \underline{2.67} \\
     &  & Random & 0.102 & 0.016 & 0.059 & 0.218 & 0.177 & 0.038 & 0.107 & 0.572 & 4.58 \\
     &  & Magnitude & 0.076 & 0.000 & 0.038 & 0.217 & 0.138 & 0.022 & 0.080 & 0.491 & 7.00 \\
     &  & STUN & \textbf{0.214} & \underline{0.027} & \underline{0.120} & \underline{0.260} & \underline{0.224} & \textbf{0.058} & \underline{0.141} & 0.649 & \textbf{2.00} \\
     &  & PruneNet & 0.121 & 0.005 & 0.063 & 0.189 & 0.157 & 0.022 & 0.089 & 0.538 & 5.75 \\
     &  & \cellcolor{aimerrow}AIMER (Ours) & \cellcolor{aimerrow}\underline{0.198} & \cellcolor{aimerrow}\textbf{0.044} & \cellcolor{aimerrow}\textbf{0.121} & \cellcolor{aimerrow}\textbf{0.263} & \cellcolor{aimerrow}0.209 & \cellcolor{aimerrow}0.040 & \cellcolor{aimerrow}0.125 & \cellcolor{aimerrow}\underline{0.672} & \cellcolor{aimerrow}\textbf{2.00} \\
    \bottomrule
  \end{tabular}%
  }
  \caption{\textbf{Expert-pruning results on DeepSeek-V2-Lite-Chat (DeepSeek), Qwen3-30B-A3B-Instruct-2507 (Qwen3), and Mixtral-8x7B-Instruct-v0.1 (Mixtral).} AIMER achieves the best rank average on DeepSeek and Qwen3 and ties for the best rank average on Mixtral, indicating a consistently strong overall capability balance across coding, creative writing, math, and multiple-choice evaluation.}
  \label{tab:main_deepseek_qwen_mixtral_summary}
\end{table*}

\begin{figure*}[!t]
  \centering
  \includegraphics[width=0.82\textwidth]{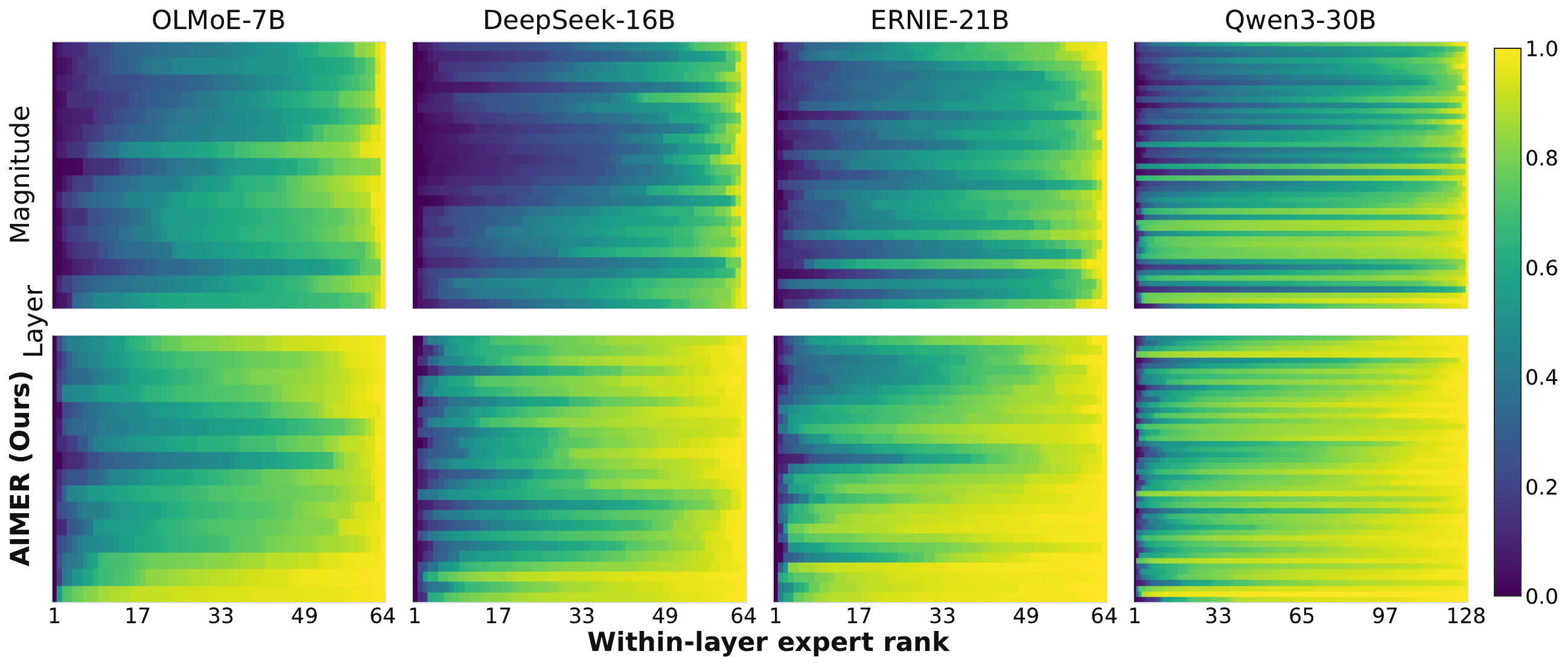}
  \caption{\textbf{Layer-wise Magnitude and AIMER score profiles across four MoE models.} Columns show OLMoE, DeepSeek, ERNIE, and Qwen3; the top and bottom rows show Magnitude and AIMER, respectively. Within each layer, experts are ranked by the corresponding score, which is min-max rescaled to $[0,1]$. Within each heatmap, experts are arranged horizontally by their within-layer rank, MoE layers are stacked vertically, and color encodes the rescaled score. Compared with Magnitude, AIMER yields more clearly stratified within-layer score profiles, making expert differences easier to distinguish.}
  \label{fig:qwen3_ranked_abs_vs_ratio_heatmap}
\end{figure*}

\begin{table*}[t]
  \centering
  \small
  \resizebox{0.75\textwidth}{!}{%
  \begin{tabular}{l c c c c c}
    \toprule
    \multirow{2}{*}{\textbf{Model}} & \multicolumn{2}{c}{\textbf{Calibration Time}} & \multicolumn{2}{c}{\textbf{Peak Calibration Memory}} & \multirow{2}{*}{\shortstack{\textbf{Loading Memory (GB)}\\\textbf{After / Before}}} \\
    \cmidrule(lr){2-3} \cmidrule(lr){4-5}
     & \textbf{REAP} & \textbf{AIMER (Ours)} & \textbf{REAP} & \textbf{AIMER (Ours)} & \\
    \midrule
    OLMoE-7B & 0.75 h & \textbf{0.22 s} & 15.51 GB & \textbf{13.00 GB} & 6.89 / 12.89 \\
    DeepSeek-16B & 1.04 h & \textbf{0.51 s} & 34.44 GB & \textbf{29.42 GB} & 15.86 / 29.27 \\
    ERNIE-21B & 1.37 h & \textbf{0.51 s} & 44.72 GB & \textbf{40.85 GB} & 21.67 / 40.66 \\
    Qwen3-30B & 2.96 h & \textbf{1.27 s} & 63.07 GB & \textbf{57.01 GB} & 29.93 / 56.92 \\
    Mixtral-47B & 1.61 h & \textbf{2.06 s} & 93.18 GB & \textbf{92.27 GB} & 44.99 / 86.99 \\
    \bottomrule
  \end{tabular}%
  }
  \caption{\textbf{Efficiency comparison between AIMER (Ours) and REAP.} All calibration measurements are collected on two NVIDIA L40S 48GB GPUs, except Mixtral, which uses four GPUs. Calibration time and peak calibration memory measure the expert-scoring stage before structural pruning. Loading memory is reported as after / before pruning at 50\% pruning ratio. Overall, AIMER incurs substantially lower resource consumption than calibration-based REAP during expert scoring.}
  \label{tab:efficiency_comparison}
\end{table*}

\subsection{Further Discussions}
\noindent \textbf{The trade-off in MoE expert pruning.} Expert pruning is fundamentally a trade-off: because experts are removed entirely, being strong on one capability can easily come with severe degradation on another, especially at higher pruning ratios. The results in \Cref{tab:main_olmoe_ernie_summary,tab:main_deepseek_qwen_mixtral_summary} show this clearly. EAN can preserve MCQ or WildBench relatively well in some settings, yet it often performs poorly on coding and math, while REAP can remain strong on math but much weaker on coding under aggressive pruning. For task-agnostic pruning, the goal is therefore not to maximize one benchmark family, but to preserve general ability. This means pruning those that are more replaceable by the remaining expert set, while retaining those that contribute more distinctive transformations. The strong balance of AIMER across coding, creative writing, math, and MCQ suggests that it is effective at identifying such replaceable experts in task-agnostic pruning.

\noindent \textbf{Magnitude versus AIMER.} Magnitude's competitive performance at moderate pruning ratios suggests that pretrained weights contain useful signals for expert pruning. AIMER's further gains indicate, however, that raw weight scale does not fully capture these signals. On Qwen3-30B at 25\% pruning, for example, it is competitive on coding and math metrics (\Cref{tab:appendix_full_wildbench_math_code}), showing that absolute weight scale can carry meaningful expert-level information. However, these gains are uneven and do not transfer reliably once the pruning decision becomes harder. At 50\% pruning on Qwen3-30B, Magnitude collapses across open-ended and reasoning benchmarks, whereas AIMER preserves substantially stronger coding, math, and multiple-choice performance (\Cref{tab:main_deepseek_qwen_mixtral_summary,tab:appendix_full_wildbench_math_code}). This suggests that raw magnitude is a useful but incomplete proxy: it captures overall expert scale, but it does not distinguish experts whose weights have similar scale but different concentration patterns. This difference is consistent with the layer-wise score profiles in \Cref{fig:qwen3_ranked_abs_vs_ratio_heatmap}. When experts are ranked by raw magnitude, many layers still contain a broad middle region of near-tied scores, leaving the pruning boundary only weakly defined. That ambiguity may be tolerable when pruning is mild, but it becomes much more costly once a larger fraction of experts is removed. AIMER, by contrast, produces a stratified within-layer score profile, shrinking this ambiguous middle region and yielding a decisive ranking.

\noindent \textbf{Resource consumption.} Since AIMER ranks experts directly from pretrained weights, it removes the expensive activation accumulation required by calibration-based methods. As shown in \Cref{tab:efficiency_comparison}, expert scoring takes only 0.22--2.06 seconds for AIMER, whereas REAP requires 0.75--2.96 hours. For all methods, the runtimes exclude model loading and checkpoint serialization and measure only the time required for expert-importance scoring. AIMER also reduces peak calibration memory across all five models. These savings in calibration time and memory become more pronounced as model scale increases. At a fixed pruning ratio, expert pruning reduces loading memory relative to the unpruned model, and the post-pruning loading memory is the same across methods.

\section{Conclusion}
In this work, we propose AIMER, a simple yet effective calibration-free criterion for task-agnostic expert pruning in MoE models. AIMER scores experts by the concentration pattern of their pretrained weights and tends to retain more functionally distinct experts after pruning. Across five MoE model families from 7B to 47B and 16 diverse benchmarks, AIMER achieves the strongest or tied-strongest overall capability balance among the evaluated expert-pruning baselines while reducing expert-scoring time from hours to seconds compared with calibration-based methods. These results show that pretrained weights alone can provide an effective signal for task-agnostic expert ranking, making calibration-free MoE pruning a practical and broadly applicable alternative to calibration-dependent expert-pruning methods.

\section*{Limitations}
This work has one main limitation: AIMER is designed for \emph{task-agnostic} expert pruning rather than task-specific compression. Since it ranks experts only from pretrained weights, it aims to identify experts that are broadly distinctive or replaceable across tasks, not experts that are most important for a particular downstream distribution. When pruning for a specific task or capability, calibration data or task-adaptive signals may still be necessary.

\section*{Acknowledgements}

This work was supported by National Key R\&D Program of China
(2024YFF0505603), the National Natural Science Foundation of China
(grant number U2541205, 62271414), Zhejiang Provincial Distinguished Young
Scientist Foundation (grant number LR23F010001), Zhejiang `Pioneer' and `Leading Goose'
R\&D Program (grant number 2024SDXHDX0006, 2024C03182), the Key Project of
Westlake Institute for Optoelectronics (grant number 2023GD007), and the 2023
International Sci-tech Cooperation Projects under the purview of the
`Innovation Yongjiang 2035' Key R\&D Program (grant number 2024Z126).

\bibliography{custom}
\appendix
\renewcommand{\dbltopfraction}{0.95}

\section{Model Architecture Details}
\label{sec:appendix_model_architecture}
Table~\ref{tab:appendix_model_architecture} summarizes the architectural diversity of the evaluated MoE models.

\section{Score Definitions of Pruning Criteria}
\label{sec:appendix_metric_scores}
For completeness, we spell out the expert-ranking scores used by the pruning criteria compared in \Cref{tab:metric_comparison}. Let $\mathcal{E}(t)$ denote the experts selected for token $t$, let $g_i(t)$ denote the router weight of expert $i$ on token $t$, and let $A_i(h_t)$ denote expert $i$'s output activation on hidden state $h_t$. We write
\[
n_i = \sum_{t:\, i \in \mathcal{E}(t)} 1
\]
for the number of tokens routed to expert $i$. 

\noindent \textbf{Frequency.}
Frequency scores an expert by how often it is selected:
\[
s_i^{\text{Freq}} = \sum_{t:\, i \in \mathcal{E}(t)} 1.
\]
This is a pure routing-frequency signal and depends only on the routing decisions collected from a calibration set.

\noindent \textbf{SEER soft counting.}
SEER replaces hard counts with the accumulated router weights:
\[
s_i^{\text{SEER}} = \sum_{t:\, i \in \mathcal{E}(t)} g_i(t).
\]
Compared with Frequency, this still depends on routing events from a calibration set, but it also uses the router confidence assigned to each selected expert.

\noindent \textbf{Expert Activation Norm (EAN).}
EAN measures an expert by the accumulated norm of its output activations:
\[
s_i^{\text{EAN}} = \sum_{t:\, i \in \mathcal{E}(t)} \|A_i(h_t)\|_2.
\]
This criterion depends on expert activations collected over the calibration tokens rather than on router weights.

\noindent \textbf{REAP.}
REAP combines router weights and activation magnitudes, then normalizes by the number of routed tokens:
\[
s_i^{\text{REAP}} = \frac{1}{n_i}\sum_{t:\, i \in \mathcal{E}(t)} g_i(t)\,\|A_i(h_t)\|_2.
\]
It therefore uses both routing information and expert activations estimated from a calibration set.

\noindent \textbf{AIMER (ours).}
AIMER is a weight-only criterion computed directly from pretrained expert parameters:
\[
s_i^{\text{AIMER}} = \frac{\|\mathbf{w}_i\|_1}{\sqrt{N_i}\,\|\mathbf{w}_i\|_2},
\]
where $\mathbf{w}_i$ is the flattened concatenation of expert $i$'s gate, up, and down projection weights, and $N_i$ is its number of parameters. Unlike the calibration-based criteria above, AIMER does not require a calibration set, collected activations, or router statistics. Instead, it compares how concentrated each expert's absolute-weight profile is. As illustrated in \Cref{fig:aimer_absmean_over_rms_example}, AIMER assigns lower scores to concentrated profiles and higher scores to diffuse profiles; experts with the largest scores are pruned within each layer.

\begin{table*}[!t]
  \centering
  \small
  \resizebox{\textwidth}{!}{%
  \begin{tabular}{l c c c c l}
    \toprule
    \textbf{Model} & \textbf{Total params} & \textbf{Experts per layer} & \textbf{Routing sparsity} & \textbf{Shared experts} & \textbf{Dense/MoE layout} \\
    \midrule
    OLMoE-1B-7B-0125-Instruct & 7B & 64 & top-8 & No & 0 dense, 16 MoE layers \\
    DeepSeek-V2-Lite-Chat & 16B & 64 routed & top-6 & Yes (2 shared) & 1 dense, 26 MoE layers \\
    ERNIE-4.5-21B-A3B-PT & 21B & 64 routed & top-6 & Yes (2 shared) & 1 dense, 27 MoE layers \\
    Qwen3-30B-A3B-Instruct-2507 & 30B & 128 & top-8 & No & 0 dense, 48 MoE layers \\
    Mixtral-8x7B-Instruct-v0.1 & 47B & 8 & top-2 & No & 0 dense, 32 MoE layers \\
    \bottomrule
  \end{tabular}%
  }
  \caption{\textbf{Architectural diversity of the evaluated MoE models.}}
  \label{tab:appendix_model_architecture}
\end{table*}

\FloatBarrier
\section{Full Benchmark Comparison}
\label{sec:appendix}
Tables~\ref{tab:appendix_full_wildbench_math_code} and~\ref{tab:appendix_full_mc} report the full benchmark comparison. We use the same method names as in the main text, and bold and underlined entries denote the best and second-best distinct pruned results within each model and pruning-ratio block.

\section{Effect of Calibration Data Choice}
\label{sec:appendix_mixing_data}
In task-agnostic pruning, the downstream task distribution is usually assumed to be unknown. We therefore use C4, a commonly adopted general-domain calibration corpus, as the primary calibration setting for a consistent comparison. Table~\ref{tab:appendix_mixing_data} further reports REAP results under C4, Evol-CodeAlpaca-v1, and Evol-CodeAlpaca-v1--C4 mixtures, with all settings using the same calibration budget of approximately 4.2M tokens and AIMER included as a calibration-free reference. The results show that, when downstream information is available, adjusting the calibration data can improve specific capabilities: adding code data substantially improves coding performance in several settings, while increasing the C4 proportion often helps recover WildBench or multiple-choice performance. However, the best trade-off varies across models and pruning ratios, and no mixture consistently performs best across capabilities. Mixed calibration is therefore a viable option when the downstream distribution is known, but it requires additional choices of calibration data and mixture proportions. AIMER focuses on the task-agnostic setting in which such information is unavailable and avoids this calibration-data selection step.

\begin{table}[!t]
  \centering
  \scriptsize
  \setlength{\tabcolsep}{2pt}
  \renewcommand{\arraystretch}{0.86}
  \resizebox{\columnwidth}{!}{%
  \begin{tabular}{l l l l | c c c}
    \toprule
    \textbf{Model} & \textbf{Ratio} & \textbf{Calibration Data} & \textbf{Method} & \textbf{Code Avg} & \textbf{WildBench} & \textbf{MC Avg} \\
    \midrule
    \multirow{10}{*}{ERNIE} & \multirow{5}{*}{25\%} & C4 & REAP & 0.126 & 0.405 & \underline{0.685} \\
     &  & CodeAlpaca & REAP & \textbf{0.534} & 0.352 & 0.665 \\
     &  & CodeAlpaca:C4 = 1:1 & REAP & \underline{0.520} & \underline{0.410} & \textbf{0.692} \\
     &  & CodeAlpaca:C4 = 1:3 & REAP & 0.501 & \textbf{0.426} & \textbf{0.692} \\
     &  & \cellcolor{aimerrow}N/A & \cellcolor{aimerrow}AIMER (Ours) & \cellcolor{aimerrow}0.455 & \cellcolor{aimerrow}0.402 & \cellcolor{aimerrow}0.681 \\
    \cmidrule(lr){2-7}
     & \multirow{5}{*}{50\%} & C4 & REAP & 0.012 & 0.199 & \underline{0.592} \\
     &  & CodeAlpaca & REAP & \textbf{0.457} & 0.204 & 0.577 \\
     &  & CodeAlpaca:C4 = 1:1 & REAP & \underline{0.434} & \underline{0.218} & 0.586 \\
     &  & CodeAlpaca:C4 = 1:3 & REAP & 0.301 & \textbf{0.236} & \textbf{0.598} \\
     &  & \cellcolor{aimerrow}N/A & \cellcolor{aimerrow}AIMER (Ours) & \cellcolor{aimerrow}0.163 & \cellcolor{aimerrow}0.217 & \cellcolor{aimerrow}\textbf{0.598} \\
    \midrule[\heavyrulewidth]
    \multirow{10}{*}{Qwen3} & \multirow{5}{*}{25\%} & C4 & REAP & 0.589 & 0.568 & 0.725 \\
     &  & CodeAlpaca & REAP & \underline{0.617} & 0.572 & 0.706 \\
     &  & CodeAlpaca:C4 = 1:1 & REAP & \textbf{0.625} & 0.591 & \underline{0.738} \\
     &  & CodeAlpaca:C4 = 1:3 & REAP & 0.602 & \textbf{0.607} & \textbf{0.740} \\
     &  & \cellcolor{aimerrow}N/A & \cellcolor{aimerrow}AIMER (Ours) & \cellcolor{aimerrow}0.595 & \cellcolor{aimerrow}\underline{0.604} & \cellcolor{aimerrow}0.723 \\
    \cmidrule(lr){2-7}
     & \multirow{5}{*}{50\%} & C4 & REAP & 0.046 & 0.226 & 0.628 \\
     &  & CodeAlpaca & REAP & \textbf{0.590} & 0.285 & 0.596 \\
     &  & CodeAlpaca:C4 = 1:1 & REAP & 0.545 & \underline{0.329} & \underline{0.666} \\
     &  & CodeAlpaca:C4 = 1:3 & REAP & \underline{0.547} & 0.301 & \textbf{0.670} \\
     &  & \cellcolor{aimerrow}N/A & \cellcolor{aimerrow}AIMER (Ours) & \cellcolor{aimerrow}0.361 & \cellcolor{aimerrow}\textbf{0.413} & \cellcolor{aimerrow}0.638 \\
    \bottomrule
  \end{tabular}%
  }
  \caption{\textbf{Effect of calibration data choice.} We report summary performance using C4, CodeAlpaca, and CodeAlpaca--C4 mixtures. All REAP settings use approximately 4.2M calibration tokens; CodeAlpaca denotes Evol-CodeAlpaca-v1.}
  \label{tab:appendix_mixing_data}
\end{table}

\begin{table*}[tbp]
  \centering
  \resizebox{\textwidth}{!}{%
  \begin{tabular}{l l l | c c c c c c c | c | c c c}
    \toprule
    \multicolumn{3}{c|}{} & \multicolumn{7}{c|}{\textbf{Code}} & \multicolumn{1}{c|}{\textbf{Creative Writing}} & \multicolumn{3}{c}{\textbf{Math}} \\
    \textbf{Model} & \textbf{Ratio} & \textbf{Method} & \textbf{HumanEval} & \textbf{HumanEval$+$} & \textbf{MBPP} & \textbf{MBPP$+$} & \textbf{Eval+} & \textbf{LiveCode} & \textbf{Code Avg} & \textbf{WildBench} & \textbf{GSM8K} & \textbf{MATH-500} & \textbf{Math Avg} \\
    \midrule
    \multirow{10}{*}{OLMoE} & 0\% & Full & 0.354 & 0.323 & 0.373 & 0.312 & 0.341 & 0.033 & 0.187 & 0.444 & 0.682 & 0.222 & 0.452 \\
    \cmidrule(lr){2-14}
       & \multirow{9}{*}{25\%} & Frequency & 0.000 & 0.000 & 0.000 & 0.000 & 0.000 & 0.000 & 0.000 & 0.127 & 0.033 & 0.024 & 0.028 \\
       &  & SEER & 0.000 & 0.000 & 0.000 & 0.000 & 0.000 & 0.000 & 0.000 & 0.141 & 0.037 & 0.012 & 0.025 \\
       &  & EAN & 0.000 & 0.000 & 0.000 & 0.000 & 0.000 & 0.000 & 0.000 & 0.205 & 0.115 & 0.006 & 0.061 \\
       &  & REAP & 0.012 & 0.012 & 0.016 & 0.016 & 0.014 & 0.000 & 0.007 & 0.260 & \underline{0.202} & \textbf{0.056} & \underline{0.129} \\
       &  & Random & \textbf{0.079} & \textbf{0.073} & \textbf{0.164} & \textbf{0.132} & \textbf{0.112} & 0.000 & \textbf{0.056} & \textbf{0.297} & 0.140 & 0.034 & 0.087 \\
       &  & Magnitude & 0.000 & 0.000 & 0.000 & 0.000 & 0.000 & 0.000 & 0.000 & 0.020 & 0.072 & 0.004 & 0.038 \\
       &  & STUN & 0.000 & 0.000 & 0.000 & 0.000 & 0.000 & 0.000 & 0.000 & 0.089 & 0.072 & 0.008 & 0.040 \\
       &  & PruneNet & \underline{0.067} & \underline{0.067} & \underline{0.140} & \textbf{0.132} & \underline{0.102} & 0.000 & \underline{0.051} & \underline{0.273} & 0.162 & \underline{0.044} & 0.103 \\
       &  & \cellcolor{aimerrow}AIMER (Ours) & \cellcolor{aimerrow}0.018 & \cellcolor{aimerrow}0.018 & \cellcolor{aimerrow}0.037 & \cellcolor{aimerrow}\underline{0.029} & \cellcolor{aimerrow}0.026 & \cellcolor{aimerrow}0.000 & \cellcolor{aimerrow}0.013 & \cellcolor{aimerrow}0.236 & \cellcolor{aimerrow}\textbf{0.221} & \cellcolor{aimerrow}\textbf{0.056} & \cellcolor{aimerrow}\textbf{0.139} \\
    \midrule[\heavyrulewidth]
    \multirow{19}{*}{ERNIE} & 0\% & Full & 0.909 & 0.878 & 0.915 & 0.765 & 0.867 & 0.247 & 0.557 & 0.479 & 0.829 & 0.780 & 0.804 \\
    \cmidrule(lr){2-14}
       & \multirow{9}{*}{25\%} & Frequency & 0.201 & 0.165 & 0.360 & 0.288 & 0.254 & 0.055 & 0.154 & 0.352 & 0.647 & 0.316 & 0.482 \\
       &  & SEER & 0.232 & 0.195 & 0.317 & 0.278 & 0.256 & 0.060 & 0.158 & 0.381 & 0.748 & 0.368 & 0.558 \\
       &  & EAN & 0.250 & 0.201 & 0.317 & 0.262 & 0.258 & 0.044 & 0.151 & 0.396 & 0.672 & 0.356 & 0.514 \\
       &  & REAP & 0.177 & 0.159 & 0.272 & 0.222 & 0.208 & 0.044 & 0.126 & \textbf{0.405} & \underline{0.782} & 0.474 & 0.628 \\
       &  & Random & 0.677 & 0.640 & 0.725 & 0.606 & 0.662 & 0.170 & 0.416 & 0.377 & 0.767 & 0.606 & 0.687 \\
       &  & Magnitude & 0.713 & 0.671 & 0.741 & 0.624 & 0.687 & \underline{0.181} & \underline{0.434} & 0.228 & 0.747 & \underline{0.640} & \underline{0.693} \\
       &  & STUN & \underline{0.744} & \underline{0.695} & \underline{0.751} & \underline{0.653} & \underline{0.711} & 0.143 & 0.427 & 0.231 & 0.662 & 0.558 & 0.610 \\
       &  & PruneNet & 0.665 & 0.646 & 0.720 & 0.601 & 0.658 & \textbf{0.192} & 0.425 & 0.381 & 0.764 & 0.610 & 0.687 \\
       &  & \cellcolor{aimerrow}AIMER (Ours) & \cellcolor{aimerrow}\textbf{0.762} & \cellcolor{aimerrow}\textbf{0.738} & \cellcolor{aimerrow}\textbf{0.767} & \cellcolor{aimerrow}\textbf{0.669} & \cellcolor{aimerrow}\textbf{0.734} & \cellcolor{aimerrow}0.176 & \cellcolor{aimerrow}\textbf{0.455} & \cellcolor{aimerrow}\underline{0.402} & \cellcolor{aimerrow}\textbf{0.830} & \cellcolor{aimerrow}\textbf{0.692} & \cellcolor{aimerrow}\textbf{0.761} \\
    \cmidrule(lr){2-14}
       & \multirow{9}{*}{50\%} & Frequency & 0.012 & 0.012 & 0.003 & 0.003 & 0.007 & 0.000 & 0.004 & 0.164 & 0.051 & 0.018 & 0.034 \\
       &  & SEER & 0.000 & 0.000 & 0.005 & 0.005 & 0.003 & 0.000 & 0.002 & 0.174 & 0.106 & 0.020 & 0.063 \\
       &  & EAN & 0.000 & 0.000 & 0.019 & 0.019 & 0.009 & 0.000 & 0.004 & \textbf{0.253} & 0.065 & 0.026 & 0.046 \\
       &  & REAP & 0.018 & 0.012 & 0.040 & 0.026 & 0.024 & 0.000 & 0.012 & 0.199 & \underline{0.567} & 0.178 & \underline{0.373} \\
       &  & Random & 0.189 & 0.165 & \underline{0.288} & 0.238 & 0.220 & 0.011 & 0.116 & 0.158 & 0.420 & \underline{0.198} & 0.309 \\
       &  & Magnitude & \underline{0.244} & \underline{0.195} & \underline{0.288} & \underline{0.243} & \underline{0.243} & \underline{0.049} & \underline{0.146} & 0.077 & 0.351 & 0.130 & 0.240 \\
       &  & STUN & 0.091 & 0.091 & 0.156 & 0.130 & 0.117 & 0.005 & 0.061 & 0.044 & 0.103 & 0.066 & 0.085 \\
       &  & PruneNet & 0.110 & 0.104 & \textbf{0.307} & \textbf{0.275} & 0.199 & 0.022 & 0.111 & 0.180 & 0.461 & 0.154 & 0.307 \\
       &  & \cellcolor{aimerrow}AIMER (Ours) & \cellcolor{aimerrow}\textbf{0.287} & \cellcolor{aimerrow}\textbf{0.250} & \cellcolor{aimerrow}0.254 & \cellcolor{aimerrow}0.225 & \cellcolor{aimerrow}\textbf{0.254} & \cellcolor{aimerrow}\textbf{0.071} & \cellcolor{aimerrow}\textbf{0.163} & \cellcolor{aimerrow}\underline{0.217} & \cellcolor{aimerrow}\textbf{0.650} & \cellcolor{aimerrow}\textbf{0.348} & \cellcolor{aimerrow}\textbf{0.499} \\
    \midrule[\heavyrulewidth]
    \multirow{10}{*}{DeepSeek} & 0\% & Full & 0.591 & 0.524 & 0.585 & 0.497 & 0.549 & 0.104 & 0.327 & 0.418 & 0.610 & 0.298 & 0.454 \\
    \cmidrule(lr){2-14}
       & \multirow{9}{*}{25\%} & Frequency & 0.000 & 0.000 & 0.000 & 0.000 & 0.000 & 0.000 & 0.000 & \underline{0.291} & 0.023 & 0.012 & 0.017 \\
       &  & SEER & 0.000 & 0.000 & 0.000 & 0.000 & 0.000 & 0.000 & 0.000 & 0.155 & 0.034 & 0.016 & 0.025 \\
       &  & EAN & 0.000 & 0.000 & 0.000 & 0.000 & 0.000 & 0.000 & 0.000 & \textbf{0.307} & \underline{0.290} & 0.018 & 0.154 \\
       &  & REAP & 0.000 & 0.000 & 0.003 & 0.003 & 0.001 & 0.000 & 0.001 & 0.152 & 0.242 & \underline{0.046} & 0.144 \\
       &  & Random & 0.000 & 0.000 & 0.000 & 0.000 & 0.000 & 0.000 & 0.000 & 0.050 & 0.081 & 0.018 & 0.050 \\
       &  & Magnitude & 0.000 & 0.000 & 0.000 & 0.000 & 0.000 & 0.000 & 0.000 & 0.003 & 0.008 & 0.002 & 0.005 \\
       &  & STUN & 0.030 & 0.030 & 0.106 & 0.093 & 0.065 & \textbf{0.011} & 0.038 & 0.137 & 0.077 & 0.030 & 0.053 \\
       &  & PruneNet & \underline{0.146} & \underline{0.122} & \textbf{0.265} & \textbf{0.209} & \textbf{0.185} & \underline{0.005} & \textbf{0.095} & 0.221 & 0.281 & 0.038 & \underline{0.160} \\
       &  & \cellcolor{aimerrow}AIMER (Ours) & \cellcolor{aimerrow}\textbf{0.152} & \cellcolor{aimerrow}\textbf{0.140} & \cellcolor{aimerrow}\underline{0.196} & \cellcolor{aimerrow}\underline{0.164} & \cellcolor{aimerrow}\underline{0.163} & \cellcolor{aimerrow}\textbf{0.011} & \cellcolor{aimerrow}\underline{0.087} & \cellcolor{aimerrow}0.246 & \cellcolor{aimerrow}\textbf{0.493} & \cellcolor{aimerrow}\textbf{0.134} & \cellcolor{aimerrow}\textbf{0.313} \\
    \midrule[\heavyrulewidth]
    \multirow{19}{*}{Qwen3} & 0\% & Full & 0.939 & 0.902 & 0.892 & 0.751 & 0.871 & 0.368 & 0.619 & 0.644 & 0.923 & 0.802 & 0.862 \\
    \cmidrule(lr){2-14}
       & \multirow{9}{*}{25\%} & Frequency & 0.000 & 0.000 & 0.000 & 0.000 & 0.000 & 0.000 & 0.000 & \textbf{0.632} & 0.904 & 0.196 & 0.550 \\
       &  & SEER & 0.006 & 0.006 & 0.000 & 0.000 & 0.003 & 0.000 & 0.002 & 0.612 & \underline{0.913} & 0.202 & 0.557 \\
       &  & EAN & 0.000 & 0.000 & 0.000 & 0.000 & 0.000 & 0.000 & 0.000 & \underline{0.626} & 0.907 & 0.200 & 0.554 \\
       &  & REAP & \textbf{0.927} & \underline{0.872} & \textbf{0.905} & \textbf{0.751} & \textbf{0.864} & \underline{0.313} & \underline{0.589} & 0.568 & \textbf{0.926} & \underline{0.778} & \textbf{0.852} \\
       &  & Random & 0.884 & 0.866 & 0.831 & 0.712 & 0.823 & 0.286 & 0.554 & 0.461 & 0.807 & 0.574 & 0.691 \\
       &  & Magnitude & 0.890 & 0.854 & \underline{0.892} & \underline{0.746} & \underline{0.845} & \underline{0.313} & 0.579 & 0.119 & 0.884 & \textbf{0.808} & \underline{0.846} \\
       &  & STUN & 0.000 & 0.000 & 0.000 & 0.000 & 0.000 & 0.000 & 0.000 & 0.001 & 0.001 & 0.004 & 0.002 \\
       &  & PruneNet & 0.854 & 0.793 & 0.817 & 0.680 & 0.786 & 0.286 & 0.536 & 0.546 & 0.616 & 0.502 & 0.559 \\
       &  & \cellcolor{aimerrow}AIMER (Ours) & \cellcolor{aimerrow}\underline{0.921} & \cellcolor{aimerrow}\textbf{0.890} & \cellcolor{aimerrow}0.862 & \cellcolor{aimerrow}0.706 & \cellcolor{aimerrow}\underline{0.845} & \cellcolor{aimerrow}\textbf{0.346} & \cellcolor{aimerrow}\textbf{0.595} & \cellcolor{aimerrow}0.604 & \cellcolor{aimerrow}0.895 & \cellcolor{aimerrow}0.756 & \cellcolor{aimerrow}0.826 \\
    \cmidrule(lr){2-14}
       & \multirow{9}{*}{50\%} & Frequency & 0.000 & 0.000 & 0.000 & 0.000 & 0.000 & 0.000 & 0.000 & 0.015 & 0.000 & 0.000 & 0.000 \\
       &  & SEER & 0.000 & 0.000 & 0.000 & 0.000 & 0.000 & 0.000 & 0.000 & 0.018 & 0.000 & 0.000 & 0.000 \\
       &  & EAN & 0.000 & 0.000 & 0.000 & 0.000 & 0.000 & 0.000 & 0.000 & \textbf{0.522} & 0.650 & 0.032 & 0.341 \\
       &  & REAP & 0.061 & 0.061 & 0.050 & 0.045 & 0.054 & 0.038 & 0.046 & 0.226 & \textbf{0.895} & \textbf{0.804} & \textbf{0.849} \\
       &  & Random & 0.006 & 0.006 & 0.013 & 0.013 & 0.010 & 0.005 & 0.007 & 0.027 & 0.085 & 0.038 & 0.061 \\
       &  & Magnitude & 0.000 & 0.000 & 0.000 & 0.000 & 0.000 & 0.000 & 0.000 & 0.000 & 0.000 & 0.000 & 0.000 \\
       &  & STUN & 0.000 & 0.000 & 0.000 & 0.000 & 0.000 & 0.000 & 0.000 & 0.000 & 0.002 & 0.006 & 0.004 \\
       &  & PruneNet & \underline{0.110} & \underline{0.104} & \underline{0.209} & \underline{0.180} & \underline{0.151} & \underline{0.055} & \underline{0.103} & 0.151 & 0.387 & 0.250 & 0.319 \\
       &  & \cellcolor{aimerrow}AIMER (Ours) & \cellcolor{aimerrow}\textbf{0.591} & \cellcolor{aimerrow}\textbf{0.549} & \cellcolor{aimerrow}\textbf{0.690} & \cellcolor{aimerrow}\textbf{0.574} & \cellcolor{aimerrow}\textbf{0.601} & \cellcolor{aimerrow}\textbf{0.121} & \cellcolor{aimerrow}\textbf{0.361} & \cellcolor{aimerrow}\underline{0.413} & \cellcolor{aimerrow}\underline{0.664} & \cellcolor{aimerrow}\underline{0.494} & \cellcolor{aimerrow}\underline{0.579} \\
    \midrule[\heavyrulewidth]
    \multirow{19}{*}{Mixtral} & 0\% & Full & 0.543 & 0.488 & 0.558 & 0.460 & 0.512 & 0.148 & 0.330 & 0.427 & 0.661 & 0.294 & 0.478 \\
    \cmidrule(lr){2-14}
       & \multirow{9}{*}{25\%} & Frequency & 0.232 & 0.189 & 0.381 & 0.307 & 0.277 & 0.038 & 0.158 & 0.356 & 0.236 & 0.058 & 0.147 \\
       &  & SEER & 0.055 & 0.055 & 0.233 & 0.206 & 0.137 & 0.000 & 0.069 & 0.135 & 0.102 & 0.012 & 0.057 \\
       &  & EAN & 0.341 & 0.287 & \underline{0.500} & \underline{0.426} & 0.388 & 0.082 & 0.235 & 0.384 & 0.475 & \underline{0.156} & 0.316 \\
       &  & REAP & \underline{0.384} & 0.329 & 0.481 & \underline{0.426} & \underline{0.405} & \underline{0.093} & \underline{0.249} & 0.397 & 0.503 & \textbf{0.176} & 0.340 \\
       &  & Random & \textbf{0.402} & \textbf{0.378} & \textbf{0.537} & \textbf{0.468} & \textbf{0.446} & 0.088 & \textbf{0.267} & \underline{0.398} & \textbf{0.541} & 0.144 & \underline{0.342} \\
       &  & Magnitude & 0.311 & 0.274 & 0.386 & 0.315 & 0.322 & 0.033 & 0.177 & 0.387 & 0.453 & 0.122 & 0.287 \\
       &  & STUN & \textbf{0.402} & \underline{0.348} & 0.458 & 0.373 & 0.395 & 0.082 & 0.239 & 0.387 & \underline{0.523} & \textbf{0.176} & \textbf{0.350} \\
       &  & PruneNet & 0.293 & 0.262 & 0.429 & 0.370 & 0.338 & 0.066 & 0.202 & 0.389 & 0.396 & 0.124 & 0.260 \\
       &  & \cellcolor{aimerrow}AIMER (Ours) & \cellcolor{aimerrow}0.366 & \cellcolor{aimerrow}0.311 & \cellcolor{aimerrow}0.474 & \cellcolor{aimerrow}0.415 & \cellcolor{aimerrow}0.391 & \cellcolor{aimerrow}\textbf{0.104} & \cellcolor{aimerrow}0.247 & \cellcolor{aimerrow}\textbf{0.431} & \cellcolor{aimerrow}0.485 & \cellcolor{aimerrow}0.148 & \cellcolor{aimerrow}0.317 \\
    \cmidrule(lr){2-14}
       & \multirow{9}{*}{50\%} & Frequency & 0.024 & 0.024 & 0.048 & 0.037 & 0.033 & 0.000 & 0.017 & 0.139 & 0.054 & 0.020 & 0.037 \\
       &  & SEER & 0.018 & 0.018 & 0.053 & 0.050 & 0.035 & 0.000 & 0.018 & 0.130 & 0.070 & 0.010 & 0.040 \\
       &  & EAN & 0.128 & 0.116 & 0.204 & 0.161 & 0.152 & 0.000 & 0.076 & 0.188 & 0.205 & 0.038 & 0.121 \\
       &  & REAP & 0.122 & 0.122 & 0.241 & 0.180 & 0.166 & 0.011 & 0.089 & 0.244 & \textbf{0.347} & \underline{0.054} & \textbf{0.201} \\
       &  & Random & 0.085 & 0.067 & 0.138 & 0.116 & 0.102 & 0.016 & 0.059 & 0.218 & 0.177 & 0.038 & 0.107 \\
       &  & Magnitude & 0.073 & 0.067 & 0.093 & 0.071 & 0.076 & 0.000 & 0.038 & 0.217 & 0.138 & 0.022 & 0.080 \\
       &  & STUN & \textbf{0.183} & \textbf{0.152} & \textbf{0.294} & \underline{0.228} & \textbf{0.214} & \underline{0.027} & \underline{0.120} & \underline{0.260} & \underline{0.224} & \textbf{0.058} & \underline{0.141} \\
       &  & PruneNet & 0.073 & 0.073 & 0.188 & 0.151 & 0.121 & 0.005 & 0.063 & 0.189 & 0.157 & 0.022 & 0.089 \\
       &  & \cellcolor{aimerrow}AIMER (Ours) & \cellcolor{aimerrow}\underline{0.140} & \cellcolor{aimerrow}\underline{0.128} & \cellcolor{aimerrow}\underline{0.288} & \cellcolor{aimerrow}\textbf{0.235} & \cellcolor{aimerrow}\underline{0.198} & \cellcolor{aimerrow}\textbf{0.044} & \cellcolor{aimerrow}\textbf{0.121} & \cellcolor{aimerrow}\textbf{0.263} & \cellcolor{aimerrow}0.209 & \cellcolor{aimerrow}0.040 & \cellcolor{aimerrow}0.125 \\
    \bottomrule
  \end{tabular}%
  }
  \caption{Full benchmark comparison on coding, creative writing, and math benchmarks.}
  \label{tab:appendix_full_wildbench_math_code}
\end{table*}

\begin{table*}[tbp]
  \centering
  \resizebox{0.85\textwidth}{!}{%
  \begin{tabular}{l l l | c c c c c c c c c}
    \toprule
    \multicolumn{3}{c|}{} & \multicolumn{9}{c}{\textbf{Multiple Choice}} \\
    \textbf{Model} & \textbf{Ratio} & \textbf{Method} & \textbf{MMLU} & \textbf{ARC-C} & \textbf{ARC-E} & \textbf{HellaSwag} & \textbf{BoolQ} & \textbf{OpenBookQA} & \textbf{RTE} & \textbf{WinoGrande} & \textbf{MC Avg} \\
    \midrule
    \multirow{10}{*}{OLMoE} & 0\% & Full & 0.534 & 0.490 & 0.758 & 0.808 & 0.766 & 0.470 & 0.711 & 0.684 & 0.653 \\
    \cmidrule(lr){2-12}
     & \multirow{9}{*}{25\%} & Frequency & 0.325 & 0.382 & 0.582 & 0.727 & 0.691 & 0.416 & 0.690 & \textbf{0.665} & 0.560 \\
     &  & SEER & 0.320 & 0.388 & 0.590 & 0.733 & 0.694 & 0.414 & \underline{0.711} & \underline{0.661} & 0.564 \\
     &  & EAN & 0.347 & 0.432 & 0.627 & \underline{0.748} & \textbf{0.734} & \underline{0.430} & 0.632 & \textbf{0.665} & 0.577 \\
     &  & REAP & 0.391 & \textbf{0.462} & \textbf{0.685} & \textbf{0.766} & 0.727 & \textbf{0.448} & 0.581 & 0.642 & \underline{0.588} \\
     &  & Random & \underline{0.432} & 0.403 & 0.598 & 0.705 & 0.706 & \underline{0.430} & 0.599 & 0.621 & 0.562 \\
     &  & Magnitude & 0.311 & 0.322 & 0.475 & 0.583 & 0.699 & 0.310 & 0.567 & 0.560 & 0.478 \\
     &  & STUN & 0.384 & 0.340 & 0.503 & 0.629 & 0.585 & 0.404 & 0.545 & 0.585 & 0.497 \\
     &  & PruneNet & 0.350 & 0.421 & 0.620 & 0.707 & 0.694 & 0.406 & 0.686 & 0.622 & 0.563 \\
     &  & \cellcolor{aimerrow}AIMER (Ours) & \cellcolor{aimerrow}\textbf{0.435} & \cellcolor{aimerrow}\underline{0.433} & \cellcolor{aimerrow}\underline{0.675} & \cellcolor{aimerrow}0.733 & \cellcolor{aimerrow}\underline{0.733} & \cellcolor{aimerrow}0.408 & \cellcolor{aimerrow}\textbf{0.758} & \cellcolor{aimerrow}0.628 & \cellcolor{aimerrow}\textbf{0.600} \\
    \midrule[\heavyrulewidth]
    \multirow{19}{*}{ERNIE} & 0\% & Full & 0.739 & 0.564 & 0.782 & 0.814 & 0.872 & 0.462 & 0.816 & 0.717 & 0.721 \\
    \cmidrule(lr){2-12}
     & \multirow{9}{*}{25\%} & Frequency & 0.571 & \underline{0.518} & 0.727 & 0.719 & 0.849 & 0.390 & 0.791 & 0.679 & 0.655 \\
     &  & SEER & 0.599 & 0.511 & 0.750 & 0.736 & 0.845 & 0.400 & 0.747 & 0.676 & 0.658 \\
     &  & EAN & 0.593 & \textbf{0.534} & \underline{0.753} & \textbf{0.789} & 0.844 & \underline{0.452} & \underline{0.805} & \textbf{0.721} & \textbf{0.687} \\
     &  & REAP & 0.643 & 0.499 & 0.717 & \underline{0.780} & 0.847 & \textbf{0.454} & \textbf{0.827} & \underline{0.711} & \underline{0.685} \\
     &  & Random & 0.643 & 0.514 & 0.732 & 0.751 & 0.833 & 0.408 & 0.751 & 0.681 & 0.664 \\
     &  & Magnitude & 0.594 & 0.507 & 0.752 & 0.699 & 0.821 & 0.416 & 0.744 & 0.654 & 0.648 \\
     &  & STUN & \underline{0.656} & 0.514 & \textbf{0.756} & 0.648 & 0.833 & 0.360 & 0.726 & 0.592 & 0.636 \\
     &  & PruneNet & \textbf{0.665} & 0.506 & 0.694 & 0.758 & \underline{0.853} & 0.448 & 0.787 & 0.706 & 0.677 \\
     &  & \cellcolor{aimerrow}AIMER (Ours) & \cellcolor{aimerrow}0.647 & \cellcolor{aimerrow}0.514 & \cellcolor{aimerrow}0.718 & \cellcolor{aimerrow}0.775 & \cellcolor{aimerrow}\textbf{0.864} & \cellcolor{aimerrow}0.430 & \cellcolor{aimerrow}\underline{0.805} & \cellcolor{aimerrow}0.694 & \cellcolor{aimerrow}0.681 \\
    \cmidrule(lr){2-12}
     & \multirow{9}{*}{50\%} & Frequency & 0.488 & 0.386 & 0.583 & 0.579 & 0.740 & 0.324 & 0.733 & 0.625 & 0.557 \\
     &  & SEER & 0.460 & 0.386 & 0.599 & 0.573 & 0.711 & 0.324 & 0.675 & 0.613 & 0.543 \\
     &  & EAN & 0.481 & \textbf{0.447} & \textbf{0.675} & \textbf{0.678} & 0.725 & \textbf{0.398} & \underline{0.744} & \textbf{0.692} & \textbf{0.605} \\
     &  & REAP & 0.479 & 0.422 & 0.621 & \underline{0.659} & 0.768 & \underline{0.392} & 0.740 & 0.651 & 0.592 \\
     &  & Random & \textbf{0.523} & \underline{0.440} & 0.630 & 0.613 & 0.767 & 0.362 & 0.733 & 0.601 & 0.584 \\
     &  & Magnitude & 0.414 & 0.357 & 0.572 & 0.508 & 0.716 & 0.324 & 0.592 & 0.543 & 0.503 \\
     &  & STUN & 0.422 & 0.337 & 0.506 & 0.476 & 0.649 & 0.326 & 0.635 & 0.542 & 0.487 \\
     &  & PruneNet & 0.493 & 0.423 & 0.607 & 0.626 & \underline{0.774} & 0.360 & \textbf{0.791} & 0.617 & 0.586 \\
     &  & \cellcolor{aimerrow}AIMER (Ours) & \cellcolor{aimerrow}\underline{0.503} & \cellcolor{aimerrow}0.426 & \cellcolor{aimerrow}\underline{0.641} & \cellcolor{aimerrow}\underline{0.659} & \cellcolor{aimerrow}\textbf{0.815} & \cellcolor{aimerrow}0.354 & \cellcolor{aimerrow}0.708 & \cellcolor{aimerrow}\underline{0.680} & \cellcolor{aimerrow}\underline{0.598} \\
    \midrule[\heavyrulewidth]
    \multirow{10}{*}{DeepSeek} & 0\% & Full & 0.567 & 0.541 & 0.785 & 0.808 & 0.829 & 0.456 & 0.726 & 0.712 & 0.678 \\
    \cmidrule(lr){2-12}
     & \multirow{9}{*}{25\%} & Frequency & 0.417 & 0.428 & 0.667 & 0.751 & 0.729 & \underline{0.424} & \textbf{0.726} & 0.671 & 0.602 \\
     &  & SEER & 0.394 & 0.451 & 0.680 & 0.754 & 0.722 & 0.422 & \textbf{0.726} & 0.669 & 0.602 \\
     &  & EAN & 0.445 & 0.477 & 0.700 & \underline{0.780} & 0.754 & \textbf{0.434} & 0.711 & \textbf{0.699} & \underline{0.625} \\
     &  & REAP & \textbf{0.485} & \textbf{0.508} & \textbf{0.756} & \textbf{0.784} & \textbf{0.771} & \textbf{0.434} & 0.693 & \underline{0.693} & \textbf{0.640} \\
     &  & Random & \underline{0.459} & 0.319 & 0.438 & 0.559 & 0.710 & 0.382 & 0.675 & 0.626 & 0.521 \\
     &  & Magnitude & 0.288 & 0.269 & 0.324 & 0.456 & 0.668 & 0.270 & 0.549 & 0.533 & 0.419 \\
     &  & STUN & 0.409 & 0.391 & 0.641 & 0.650 & 0.717 & 0.366 & 0.625 & 0.635 & 0.554 \\
     &  & PruneNet & 0.438 & 0.451 & 0.673 & 0.723 & \textbf{0.771} & 0.400 & 0.708 & 0.669 & 0.604 \\
     &  & \cellcolor{aimerrow}AIMER (Ours) & \cellcolor{aimerrow}0.454 & \cellcolor{aimerrow}\underline{0.481} & \cellcolor{aimerrow}\underline{0.711} & \cellcolor{aimerrow}0.749 & \cellcolor{aimerrow}\underline{0.761} & \cellcolor{aimerrow}0.412 & \cellcolor{aimerrow}\underline{0.718} & \cellcolor{aimerrow}0.683 & \cellcolor{aimerrow}0.621 \\
    \midrule[\heavyrulewidth]
    \multirow{19}{*}{Qwen3} & 0\% & Full & 0.802 & 0.625 & 0.838 & 0.797 & 0.887 & 0.446 & 0.769 & 0.736 & 0.737 \\
    \cmidrule(lr){2-12}
     & \multirow{9}{*}{25\%} & Frequency & \underline{0.761} & 0.625 & 0.839 & \underline{0.795} & 0.886 & 0.442 & 0.773 & \underline{0.732} & 0.732 \\
     &  & SEER & 0.760 & 0.630 & 0.845 & \underline{0.795} & \textbf{0.888} & \textbf{0.446} & 0.776 & \underline{0.732} & \underline{0.734} \\
     &  & EAN & \textbf{0.765} & \underline{0.631} & \textbf{0.849} & \textbf{0.796} & \underline{0.887} & \underline{0.444} & 0.780 & \textbf{0.736} & \textbf{0.736} \\
     &  & REAP & 0.733 & \textbf{0.637} & \underline{0.848} & 0.766 & 0.871 & \textbf{0.446} & 0.780 & 0.718 & 0.725 \\
     &  & Random & 0.717 & 0.531 & 0.763 & 0.707 & 0.883 & 0.402 & 0.762 & 0.707 & 0.684 \\
     &  & Magnitude & 0.714 & 0.560 & 0.803 & 0.665 & 0.859 & 0.398 & 0.718 & 0.642 & 0.670 \\
     &  & STUN & 0.240 & 0.247 & 0.263 & 0.270 & 0.508 & 0.234 & 0.520 & 0.506 & 0.348 \\
     &  & PruneNet & 0.721 & 0.573 & 0.813 & 0.716 & 0.874 & 0.402 & \underline{0.783} & 0.714 & 0.700 \\
     &  & \cellcolor{aimerrow}AIMER (Ours) & \cellcolor{aimerrow}0.715 & \cellcolor{aimerrow}0.615 & \cellcolor{aimerrow}0.842 & \cellcolor{aimerrow}0.772 & \cellcolor{aimerrow}0.886 & \cellcolor{aimerrow}0.428 & \cellcolor{aimerrow}\textbf{0.794} & \cellcolor{aimerrow}\underline{0.732} & \cellcolor{aimerrow}0.723 \\
    \cmidrule(lr){2-12}
     & \multirow{9}{*}{50\%} & Frequency & 0.276 & 0.287 & 0.391 & 0.436 & 0.655 & 0.306 & 0.585 & 0.564 & 0.437 \\
     &  & SEER & 0.276 & 0.288 & 0.396 & 0.435 & 0.651 & 0.302 & 0.567 & 0.571 & 0.436 \\
     &  & EAN & \textbf{0.634} & \textbf{0.544} & \textbf{0.770} & \textbf{0.786} & \textbf{0.885} & \textbf{0.440} & \textbf{0.783} & \textbf{0.731} & \textbf{0.697} \\
     &  & REAP & 0.532 & \underline{0.537} & \underline{0.734} & 0.615 & 0.836 & 0.370 & 0.729 & 0.674 & 0.628 \\
     &  & Random & 0.238 & 0.280 & 0.365 & 0.393 & 0.649 & 0.262 & 0.581 & 0.501 & 0.409 \\
     &  & Magnitude & 0.263 & 0.238 & 0.328 & 0.291 & 0.514 & 0.302 & 0.516 & 0.527 & 0.372 \\
     &  & STUN & 0.239 & 0.230 & 0.271 & 0.265 & 0.501 & 0.236 & 0.484 & 0.508 & 0.342 \\
     &  & PruneNet & \underline{0.580} & 0.451 & 0.707 & 0.573 & 0.696 & 0.348 & \underline{0.780} & 0.622 & 0.595 \\
     &  & \cellcolor{aimerrow}AIMER (Ours) & \cellcolor{aimerrow}0.542 & \cellcolor{aimerrow}0.482 & \cellcolor{aimerrow}0.724 & \cellcolor{aimerrow}\underline{0.653} & \cellcolor{aimerrow}\underline{0.847} & \cellcolor{aimerrow}\underline{0.378} & \cellcolor{aimerrow}0.769 & \cellcolor{aimerrow}\underline{0.712} & \cellcolor{aimerrow}\underline{0.638} \\
    \midrule[\heavyrulewidth]
    \multirow{19}{*}{Mixtral} & 0\% & Full & 0.691 & 0.649 & 0.841 & 0.862 & 0.887 & 0.500 & 0.726 & 0.742 & 0.737 \\
    \cmidrule(lr){2-12}
     & \multirow{9}{*}{25\%} & Frequency & 0.403 & 0.467 & 0.676 & 0.759 & 0.804 & 0.424 & 0.625 & 0.592 & 0.594 \\
     &  & SEER & 0.260 & 0.317 & 0.467 & 0.626 & 0.675 & 0.328 & 0.531 & 0.522 & 0.466 \\
     &  & EAN & 0.614 & \textbf{0.610} & 0.820 & \underline{0.840} & 0.871 & \textbf{0.494} & 0.686 & \underline{0.732} & 0.708 \\
     &  & REAP & \underline{0.630} & \underline{0.609} & \textbf{0.823} & 0.835 & \underline{0.875} & 0.482 & \textbf{0.765} & 0.713 & \textbf{0.717} \\
     &  & Random & 0.620 & 0.603 & 0.798 & 0.829 & 0.859 & 0.482 & 0.704 & 0.725 & 0.702 \\
     &  & Magnitude & 0.518 & 0.495 & 0.724 & 0.763 & 0.803 & 0.420 & 0.650 & 0.634 & 0.626 \\
     &  & STUN & \textbf{0.639} & \textbf{0.610} & 0.795 & 0.831 & 0.869 & \underline{0.488} & \underline{0.740} & 0.729 & \underline{0.713} \\
     &  & PruneNet & 0.473 & 0.490 & 0.728 & 0.751 & 0.800 & 0.410 & 0.625 & 0.592 & 0.609 \\
     &  & \cellcolor{aimerrow}AIMER (Ours) & \cellcolor{aimerrow}0.627 & \cellcolor{aimerrow}0.603 & \cellcolor{aimerrow}\underline{0.822} & \cellcolor{aimerrow}\textbf{0.843} & \cellcolor{aimerrow}\textbf{0.876} & \cellcolor{aimerrow}0.486 & \cellcolor{aimerrow}0.671 & \cellcolor{aimerrow}\textbf{0.738} & \cellcolor{aimerrow}0.708 \\
    \cmidrule(lr){2-12}
     & \multirow{9}{*}{50\%} & Frequency & 0.323 & 0.381 & 0.572 & 0.662 & 0.745 & 0.366 & 0.549 & 0.566 & 0.520 \\
     &  & SEER & 0.329 & 0.381 & 0.600 & 0.678 & 0.720 & 0.366 & 0.541 & 0.570 & 0.523 \\
     &  & EAN & \underline{0.530} & \underline{0.550} & 0.754 & \textbf{0.802} & 0.853 & \textbf{0.458} & \underline{0.737} & \underline{0.707} & \textbf{0.674} \\
     &  & REAP & \textbf{0.542} & 0.527 & \underline{0.758} & 0.791 & \underline{0.861} & 0.444 & \textbf{0.751} & 0.688 & 0.670 \\
     &  & Random & 0.407 & 0.414 & 0.634 & 0.693 & 0.782 & 0.402 & 0.639 & 0.601 & 0.572 \\
     &  & Magnitude & 0.321 & 0.355 & 0.548 & 0.577 & 0.683 & 0.354 & 0.549 & 0.542 & 0.491 \\
     &  & STUN & 0.522 & 0.503 & 0.743 & 0.760 & 0.819 & 0.438 & 0.722 & 0.688 & 0.649 \\
     &  & PruneNet & 0.367 & 0.396 & 0.619 & 0.653 & 0.756 & 0.352 & 0.607 & 0.558 & 0.538 \\
     &  & \cellcolor{aimerrow}AIMER (Ours) & \cellcolor{aimerrow}0.524 & \cellcolor{aimerrow}\textbf{0.567} & \cellcolor{aimerrow}\textbf{0.779} & \cellcolor{aimerrow}\underline{0.795} & \cellcolor{aimerrow}\textbf{0.867} & \cellcolor{aimerrow}\underline{0.446} & \cellcolor{aimerrow}0.693 & \cellcolor{aimerrow}\textbf{0.709} & \cellcolor{aimerrow}\underline{0.672} \\
    \bottomrule
  \end{tabular}%
  }
  \caption{Full benchmark comparison on multiple-choice benchmarks.}
  \label{tab:appendix_full_mc}
\end{table*}

\section{Code-Domain CKA Analysis}
\label{sec:appendix_code_cka}
\begin{figure*}[t]
  \centering
  \includegraphics[width=\textwidth]{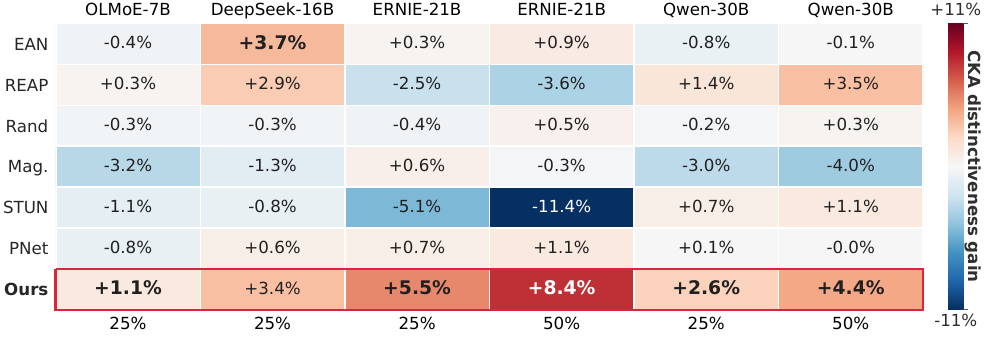}
  \caption{\textbf{Change in expert distinctiveness after pruning on code-domain inputs.} We repeat the analysis in \Cref{fig:cka_distinctness_gain} using 4096 tokens from Evol-CodeAlpaca-v1 while keeping the expert subsets fixed. Each cell reports the layer-averaged relative decrease in mean CKA after pruning; positive values indicate that the retained experts are more functionally distinct. AIMER yields positive gains in all six settings and is strongest in five of them.}
  \label{fig:cka_distinctness_gain_code}
\end{figure*}

To test whether the CKA trend in \Cref{fig:cka_distinctness_gain} is specific to C4 inputs, we repeat the analysis using 4096 tokens from Evol-CodeAlpaca-v1~\cite{luo2023wizardcoder}. We keep the selected expert subsets unchanged: calibration-based methods use the same C4-selected subsets as in \Cref{fig:cka_distinctness_gain}, while AIMER uses the same calibration-free subsets. The code-domain inputs are therefore used only for post-hoc CKA evaluation and do not affect expert ranking or pruning. As shown in \Cref{fig:cka_distinctness_gain_code}, AIMER produces positive distinctiveness gains in all six settings and is strongest in five of them. This result is consistent with the C4 analysis and provides further evidence that AIMER retains relatively distinct experts under a domain shift to code.

\section{Radar Plots by Benchmark}
\label{sec:appendix_radar}

\begin{figure*}[t]
  \centering
  \includegraphics[width=\textwidth]{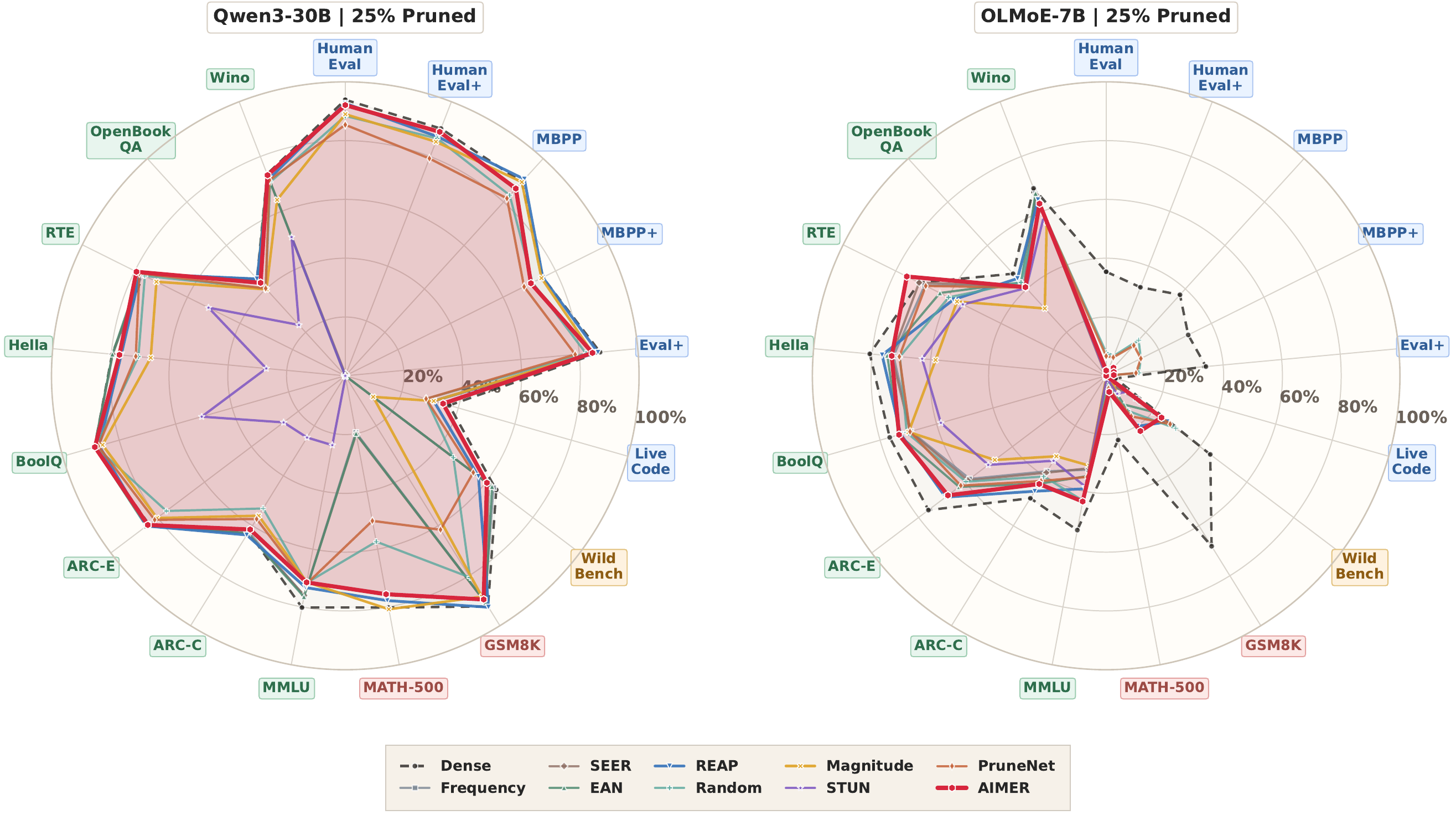}
  \caption{\textbf{Radar plots of Qwen3-30B and OLMoE-7B performance across all benchmarks at 25\% pruning.} The left and right panels show Qwen3-30B and OLMoE-7B, respectively.}
  \label{fig:appendix_radar_olmoe_qwen25}
\end{figure*}

\begin{figure*}[t]
  \centering
  \includegraphics[width=\textwidth]{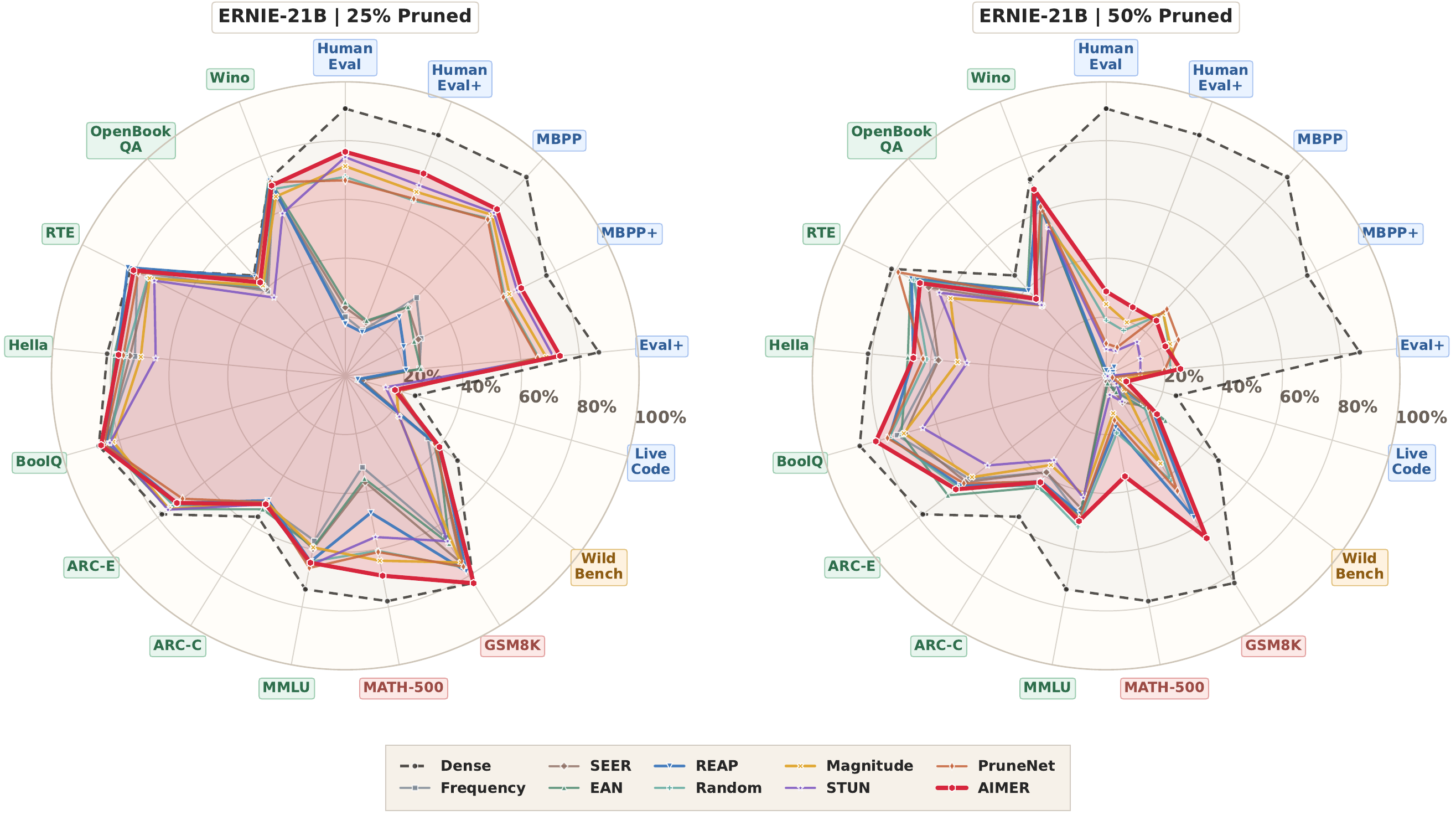}
  \caption{\textbf{Radar plots of ERNIE-21B performance across all benchmarks.} The left and right panels show 25\% and 50\% pruning ratios, respectively.}
  \label{fig:appendix_radar_ernie}
\end{figure*}

\begin{figure*}[t]
  \centering
  \includegraphics[width=\textwidth]{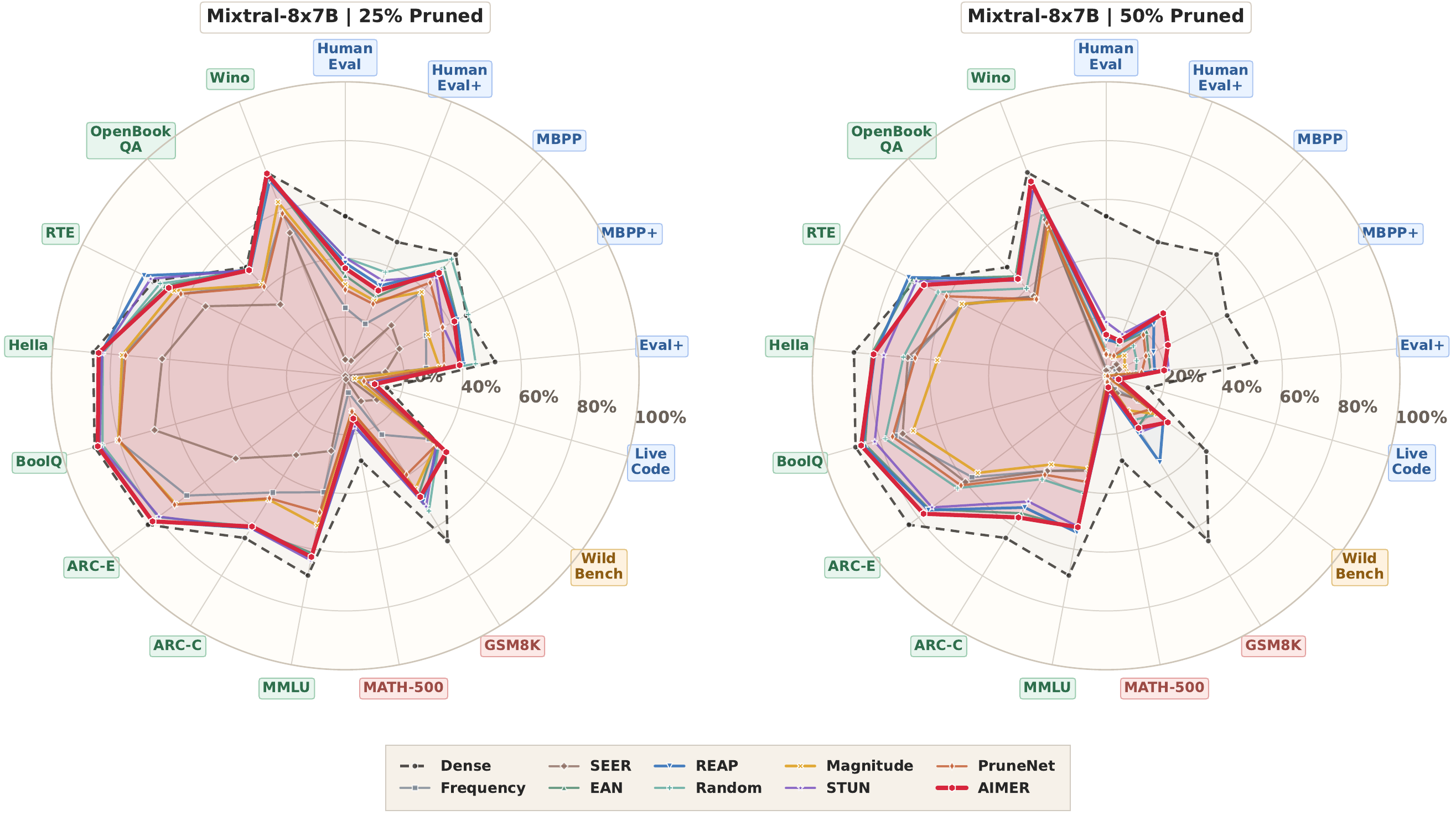}
  \caption{\textbf{Radar plots of Mixtral performance across all benchmarks.} The left and right panels show 25\% and 50\% pruning ratios, respectively.}
  \label{fig:appendix_radar_mixtral}
\end{figure*}

\begin{figure}[!t]
  \centering
  \includegraphics[width=\columnwidth]{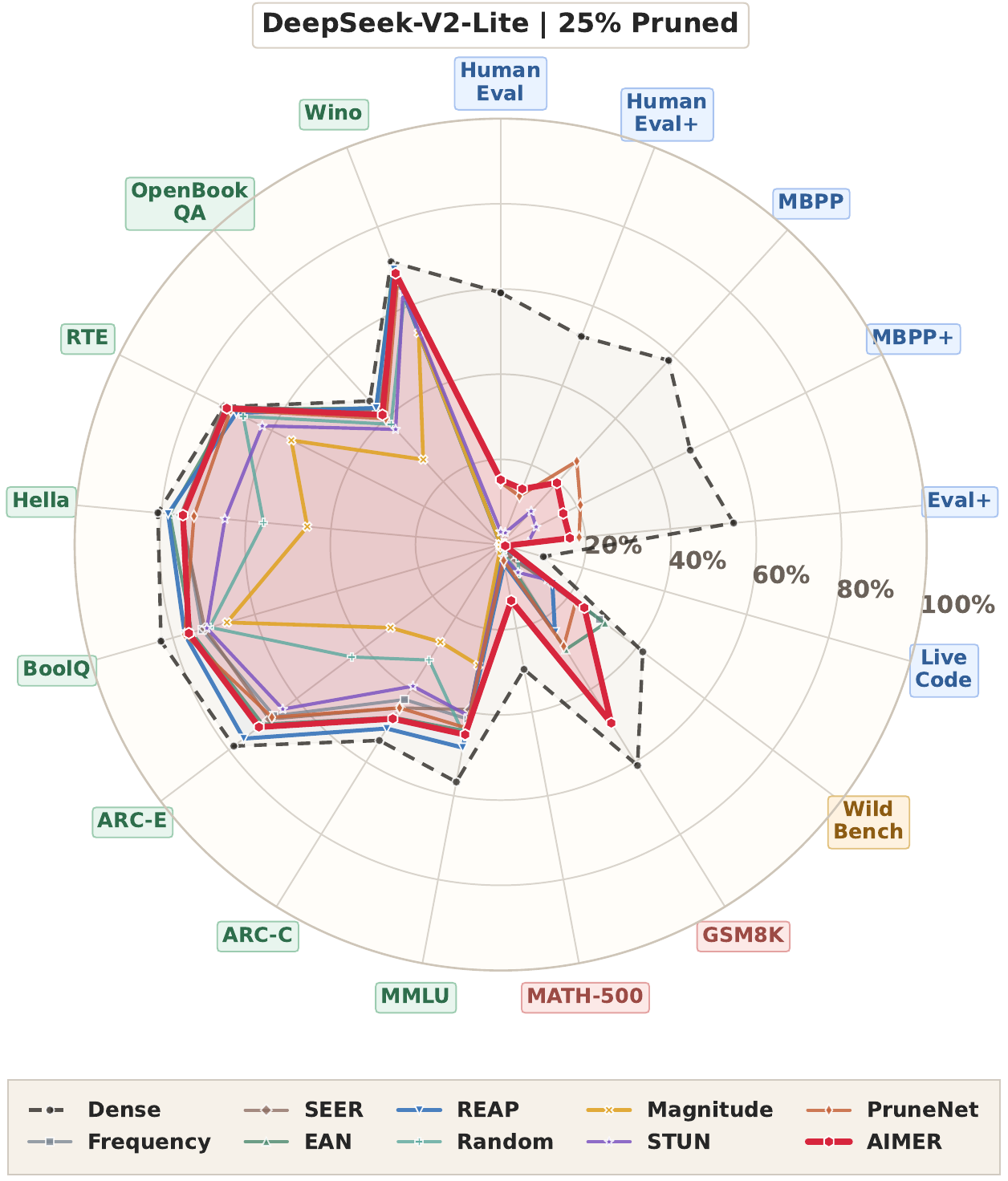}
  \caption{\textbf{Radar plot of DeepSeek performance across all benchmarks at 25\% pruning.}}
  \label{fig:appendix_radar_deepseek}
\end{figure}

Figures~\ref{fig:appendix_radar_olmoe_qwen25}, \ref{fig:appendix_radar_ernie}, \ref{fig:appendix_radar_mixtral}, and \ref{fig:appendix_radar_deepseek} visualize the per-benchmark trade-offs behind the appendix tables. The dashed outline denotes the dense model, and each colored trace denotes a pruning method. Across models, AIMER contracts the dense-model contour more uniformly, while several baselines collapse on one or more capability clusters.

\makeatletter
\setlength{\@fptop}{0pt}
\makeatother

\end{document}